\pdfoutput=1
\documentclass[11pt]{article}

\usepackage{emnlp2021}

\usepackage{times}
\usepackage{latexsym}

\usepackage[T1]{fontenc}
\usepackage[utf8]{inputenc}

\usepackage{microtype}

\usepackage[flushleft]{threeparttable}
\usepackage{todonotes}
\usepackage{enumitem}
\usepackage{algorithm}
\usepackage{algpseudocode}
\usepackage{booktabs}
\usepackage{subcaption}
\usepackage[nointegrals]{wasysym}
\usepackage{tikz}
\usepackage{tabu}
\usepackage{tabularx}
\usepackage{float}
\usepackage{listings}
\usepackage{color}
\usepackage{multirow}
\usepackage{colortbl}
\usepackage{framed}
\usepackage{pifont}
\usepackage{dsfont}
\usepackage{xcolor}
\usepackage{amsmath,amsthm}
\usepackage{amssymb}
\usepackage{graphicx}
\usepackage{textcomp}
\usepackage{tcolorbox}
\usepackage{soul}
\usepackage{lipsum}
\definecolor{DarkRed}{RGB}{130,25,0}

\newcommand{\dataset}{\emph{ESTER}}

\newcommand{\red}{\textsc{RED}}
\newcommand{\causal}{\textsc{Causal}}
\newcommand{\cause}{\textsc{CAUSE}}
\newcommand{\enable}{\textsc{ENABLE}}
\newcommand{\prevent}{\textsc{PREVENT}}
\newcommand{\subevent}{\textsc{Sub-event}}
\newcommand{\coref}{\textsc{Co-reference}}

\newcommand{\condition}{\textsc{Conditional}}
\newcommand{\counter}{\textsc{Counterfactual}}

\newcommand{\hieve}{\textsc{HiEve}}

\newcommand{\ignore}[1]{}
\definecolor{mypurple}{RGB}{121,55,196}
\definecolor{myblue}{RGB}{60,177,245}
\definecolor{myorange}{RGB}{243,206,84}
\definecolor{mygreen}{RGB}{41,222,35}
\definecolor{light-gray}{gray}{0.9}

\newcounter{exctr}

\newcounter{eventCtr}

\newcounter{timexCtr}

\newcommand{\mypar}[1]{\noindent{\textbf{#1\ }}}
\usepackage{titlesec}
\titlespacing*{\section}{0pt}{*0.97}{*0.95}
\titlespacing*{\subsection}{0pt}{*0.91}{*0.90}
\title{ESTER: A Machine Reading Comprehension Dataset for Reasoning about Event Semantic Relations}

\author{Rujun Han$^{1}$ ~ I-Hung Hsu$^{1}$ ~ Jiao Sun$^{1}$ ~ Julia Baylon$^{2}$\\
{\bf Qiang Ning$^{3}$\thanks{\hspace{0.2cm}Part of the work was done while the author was at the Allen Institute for AI.} ~ Dan Roth$^{4}$ ~ Nanyun Peng$^{1,2}$}\\
$^1$University of Southern California ~ $^2$University of California, Los Angeles \\
$^3$Amazon ~ $^4$University of Pennsylvania \\
{\tt \{rujunhan,ihunghsu,jiaosun\}@usc.edu; juliabaylon@ucla.edu} \\
{\tt qning@amazon.com;  danroth@seas.upenn.edu} \\ {\tt violetpeng@cs.ucla.edu}
}

\begin{document}
\maketitle

\begin{abstract}
Understanding how events are semantically related to each other is the essence of reading comprehension. Recent event-centric reading comprehension datasets focus mostly on event arguments or temporal relations. While these tasks partially evaluate machines' ability of narrative understanding, human-like reading comprehension requires the capability to process event-based information beyond arguments and temporal reasoning. For example, to understand causality between events, we need to infer motivation or purpose; to establish event hierarchy, we need to understand the composition of events. To facilitate these tasks, we introduce {\dataset}, a comprehensive machine reading comprehension (MRC) dataset for \textbf{E}vent \textbf{S}eman\textbf{t}ic R\textbf{e}lation \textbf{R}easoning. The dataset leverages natural language queries to reason about the five most common event semantic relations, provides more than 6K questions, and captures 10.1K event relation pairs. Experimental results show that the current SOTA systems achieve 22.1\%, 63.3\% and 83.5\% for token-based exact-match (\textbf{EM}), $F_1$ and event-based \textbf{HIT@1} scores, which are all significantly below human performances (36.0\%, 79.6\%, 100\% respectively), highlighting our dataset as a challenging benchmark. \footnote{Data, models and reproduction code are available here: https://github.com/PlusLabNLP/ESTER.}
\end{abstract}
\section{Introduction}
\label{sec:intro}
\begin{figure}[t]
    \centering
\includegraphics[trim=0.2cm 0.8cm 1.2cm 0cm, clip, width=0.99\columnwidth]{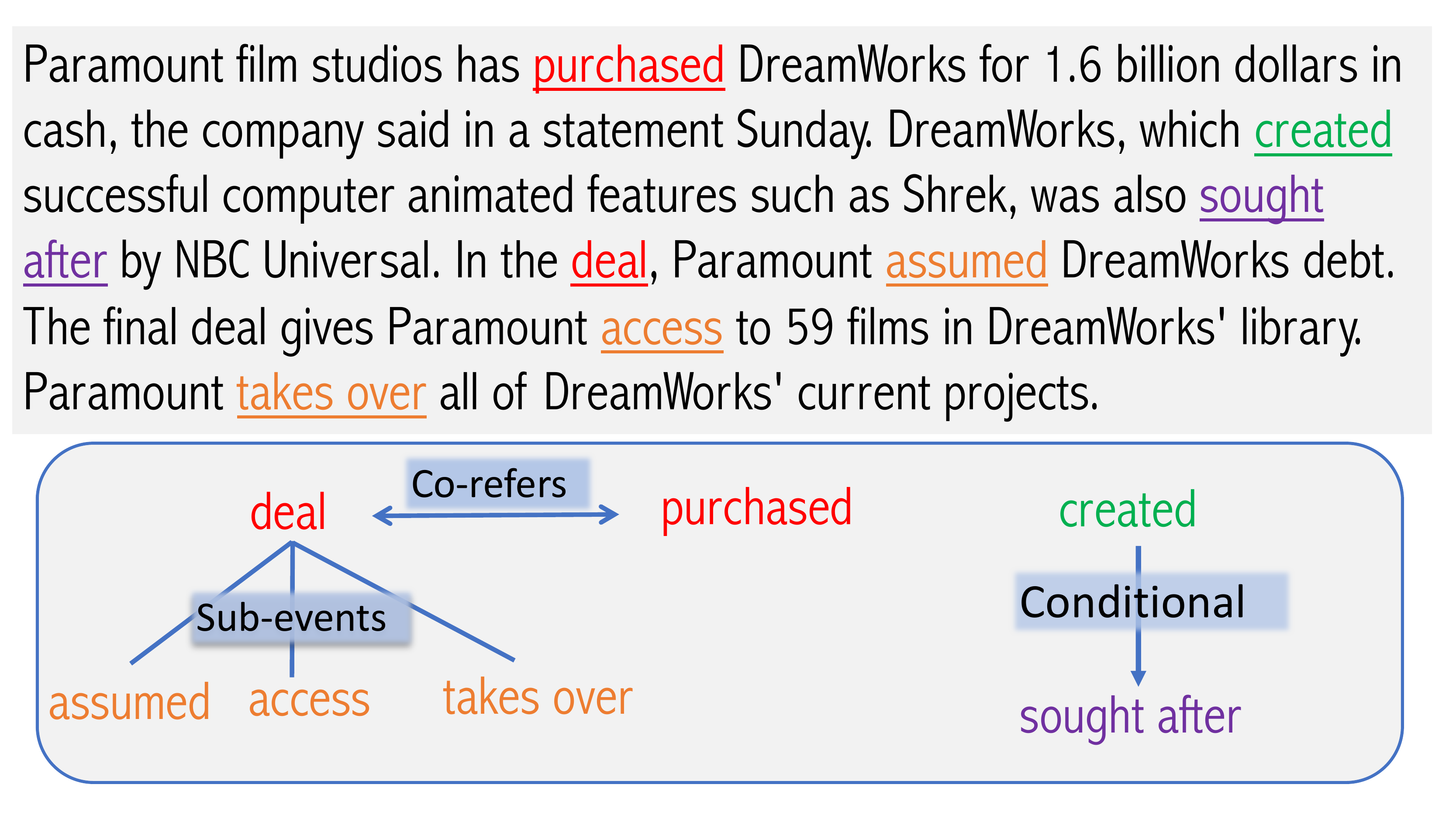}
\vspace{-0.5cm}
\caption{A graph illustration of event semantic relations in narratives. We use trigger words to represent events in this graph. }
\label{fig:illustrating-exp}
\vspace{-0.5cm}
\end{figure}

%\hl{what's event semantic relation and why is it important}
Narratives such as stories and news articles are composed of series of events~\cite{careyNsnodgrass, harmonhandbook}. 
Understanding how events are logically connected is essential for reading comprehension \cite{caselli-vossen-2017-event, mostafazadeh-etal-2016-caters}. For example, Figure~\ref{fig:illustrating-exp} illustrates several pairwise relations for events in the given passage: \textit{``the deal''} can be considered as the same event of \textit{``Paramount purchased DreamWorks,''} forming a coreference relation; it is also a complex event that contains \textit{``assumed debt,'' ``gives access''} and \textit{``takes over projects''} as its sub-events. The event \textit{``sought after''} is facilitated by a previous event \textit{``created features.''} By capturing these event semantic relations, people can often grasp the gist of a story. Therefore, for machines to achieve human-level narrative understanding, we need to test and ensure models' capability to reason over these event relations. 
%\IHNote{I'm not sure whether we need to define our definition of event first here. Like we use trigger words to represent the event etc.}
% \violets{what does it mean by `is conditional on'? should use more natural language to describe such relation.}

\begin{figure*}[t]
    \centering
\includegraphics[trim=5.4cm 0cm 5.4cm 0cm, clip, width=0.72\columnwidth,angle=-90]{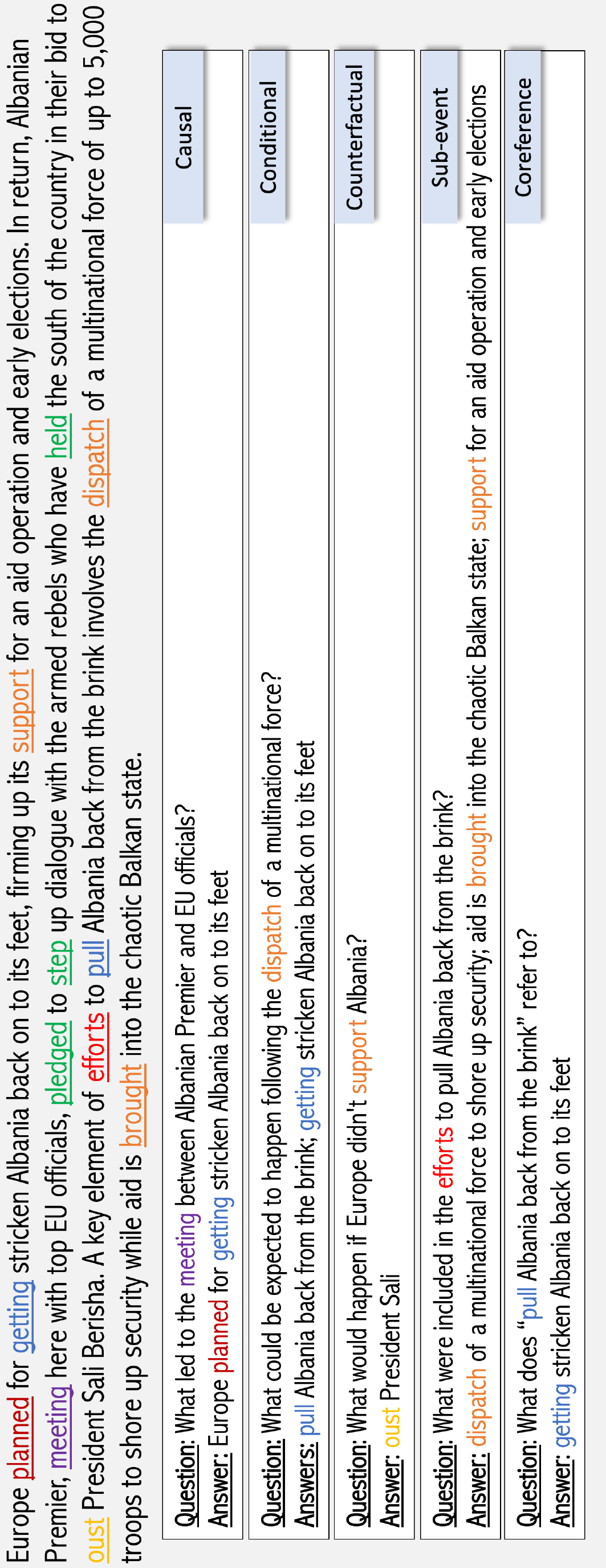}
\vspace{-0.3cm}
\caption{Examples of event annotations and 5 types of QAs in our dataset. Not all  events are annotated for clarity purpose. Different colors are used for better visualization.} %\IHNote{Any logic meanings for different colored arrangement of event trigger?}}
\label{fig:mrc-exp}
\vspace{-0.6cm}
\end{figure*}

%\hl{what's MRC for event semantic relation}
% \violets{I move this forward since I think it's better to have an overview in the beginning. Then we can break it down to talk about individual feature of this dataset.}
% \violets{low?}\ihung{low inter-annotator "AGREEMENT"}
In this work, we study five types of \textbf{event semantic relations}: \causal, \subevent, \coref, {\condition} and {\counter}, and propose to use natural language questions to reason about \textbf{event semantic relations}. Figure~\ref{fig:mrc-exp} shows example question-answer pairs for each relation type. 

Although previous works studied some subset of these relations such as {\subevent} \cite{glavas-etal-2014-hieve, yao-etal-2020-weakly}, {\causal} and {\condition} \cite{mirza-etal-2014-annotating, mirza-tonelli-2014-analysis, ogorman-etal-2016-richer}, most of them adopted pairwise relation extraction (RE) formulation by constructing (event, event, relation) triplets and predicting the relation for the pair of events. Event relations of RE formulation are rigidly defined as class labels based on expert knowledge, which could suffer from relatively low inter-annotator agreements \cite{glavas-etal-2014-hieve, ogorman-etal-2016-richer} and may not be the most natural way to exploit the semantic connections between relations and events in their context.

We instead propose to reason about \textit{event semantic relations} as a \textit{reading comprehension / question answering} task. Natural language queries ease the annotation efforts in the RE formulation by supplementing expert-defined relations with textual prompts. When querying {\causal} relations, we can ask \textit{``what causes / leads to Event A?''} or \textit{``why does A happen?''}; when reasoning {\subevent} relation, we can ask \textit{``what are included in Event B?''} or \textit{``What does B entail?''} etc. 
\textit{``lead to,''} \textit{``included''} and \textit{``entail,''} as textual cues, can help models better understand which relation is being queried.

% Moreover, the flexibility enabled by natural language queries allows us to define events using both triggers and arguments such as subject, object, time, and location.\violets{it seems an RE dataset can also define and annotate all the argument and predict relation based on all the information, right? Are you saying all prior works didn't do this, or the formulation does not support this? We need to be super clear.} For example, we can easily distinguish \textit{``Paramount purchases DreamWorks''} from \textit{``NBC Universal purchases DreamWorks''} based on different subjects for the same trigger word \textit{``purchases.''} Similarly, other events arguments can be naturally incorporated in questions to help disambiguate events. This improvement is particularly helpful when multiple identical or similar triggers with different arguments exist in the context. 

%\hl{why MRC for event semantic relation}
Our question-answering task also poses unique challenges for reasoning \textit{event semantic relations}. First, the correct answers can be completely different with slight changes of queries. In Figure~\ref{fig:mrc-exp}, if we modify the third question to be \textit{``What would happen if Europe supported Albania?''} then \textit{``oust President Sali''} becomes an invalid answer. This challenge allows us to test whether models possess robust reasoning skills or simply conduct pattern matching. Second, answers must be in the form of complete and meaningful text spans. For the {\counter} example in Figure~\ref{fig:mrc-exp}, a random text span \textit{``President Sali Berisha''} is not a meaningful answer while a shortened answer \textit{``oust''} is not complete. To get correct answers, models need to detect both event triggers and their event arguments. Finally, there could be multiple valid events in a passage that can answer a question, and a good system should be able to identify different valid answers simultaneously as in the {\subevent} QA of Figure~\ref{fig:mrc-exp}. These challenges make our task more difficult than the classification tasks in RE.
% \violets{probably also add the argument that one question could have multiple answers.} 
%\violets{point to an example (figure 2 subevent could do)}

%\IHNote{multi-class?? or multi-label, I think you mean multi-class}
%\IHNote{"Event Extraction as Machine Reading Comprehension" by Liu et al. do similar things.}
%\hl{contrast with previous related work and weakness}
A few noticeable event-centric MRC datasets have been proposed recently. TORQUE \cite{ning-etal-2020-torque} and MCTACO \cite{zhou-etal-2019-going} are two recent MRC datasets that study event temporal relations. However, knowing only the temporal aspect of events could not solve many important event semantic relations. For example, in Figure~\ref{fig:illustrating-exp}, to understand that \textit{``assumed debt,'' ``gives access''} and \textit{``takes over projects''} are sub-events of \textit{``the deal,''} a model not only needs to know that all these four events have overlapped time intervals but also share the same associated participants for \textit{``the deal''} to contain the other three.

% \violets{I'll also say that this is the first comprehensive dataset that contains all the/several significant semantic types, which helps us understand their interplay. Maybe say this before you discuss the problem formulation since later you spend more time discussing problem formulation.}

% \violets{here you might want to re-structure this into a contribution list, since many of the points have been made earlier on in the paper as well (e.g., the MRC format, 5 event semantic relations, natural questions, requires complete answers, etc.), so it's more like a summary of the previous paragraphs with a little bit new information. Probably more natural as a contribution list that way.}
We summarize our contributions below. 
    \vspace{-0.4cm}
\begin{enumerate}
    \item We introduce \dataset, the first comprehensive MRC / QA dataset for the five proposed \textbf{event semantic relations} by adopting natural language questions and requiring complete event spans in the passage as answers.
    \vspace{-0.4cm}
    \item By proposing a generative QA task that models all five relations jointly and comparing it with traditional extractive QA task, we provide insights on how these event semantic relations interplay for MRC.
    \vspace{-0.4cm}
    \item Our experimental results reveal SOTA models' deficiencies in our target tasks, which demonstrates that {\dataset} is a challenging dataset that can facilitate future research in MRC for \textbf{event semantic relations}.
\end{enumerate}

\section{Definitions}
\label{sec:define}
Composing event-centric questions and answers requires identifications of both events and their relations. In this section, we describe our definitions of events and five \textbf{event semantic relations}.

\subsection{Events}

Adopting the general guideline of \citet{ACE}, we define an event as a trigger word and its arguments (subject, object, time and location). An event trigger is a word that most clearly describes the event's occurrence, and it is often a verb or noun that evokes the action or the status of the target event \cite{TimeBank}. Later event-centric reasoning work mostly uses this trigger definition, e.g., TE3 \citep{uzzaman-etal-2013-semeval}, HiEve \citep{glavas-etal-2014-hieve}, RED \citep{ogorman-etal-2016-richer} and TORQUE \citep{ning-etal-2020-torque}.

While event triggers must exist in the context, some event arguments need to be inferred by annotators. In Figure~\ref{fig:mrc-exp}, for example, \textit{``getting''} is an event trigger and its subject, object and location are \textit{``Europe,''} \textit{``Albania''} and \textit{``Europe''} respectively. The event's time can be inferred to be approximately the document writing time. To ensure event-centric reasoning, we require all questions and answers to include a trigger. Annotators are allowed to use any event arguments including those inferred to make questions natural and descriptive; whereas for answers, they need to identify complete and meaningful text spans in the passage.

\subsection{Event Semantic Relations}

Next, we discuss the definitions of the five types of event semantic relations in our dataset, most of which are consistent with previous studies. For example, {\causal} and {\condition} have been studied in \citet{Wolff-2007, do-etal-2011-minimally, mirza-tonelli-2014-analysis, mirza-etal-2014-annotating}. {\subevent} and {\coref} were studied in \citet{glavas-etal-2014-hieve, ogorman-etal-2016-richer}. Cosmos QA \cite{huang-etal-2019-cosmos} has a small amount of {\counter} questions, but it is not an event-centric dataset. The examples we use below are all presented in Figure~\ref{fig:mrc-exp}.

\paragraph{Causal:} A pair of events ($e_i, e_j$) exhibits a {\causal} relation \textbf{if $e_i$ happens then $e_j$ will definitely happen} according to the given passage. For example, the passage explicitly says that the \textit{``meeting''} happens \textit{``in return''} for \textit{``Europe planned for getting stricken Albanian back.''} Therefore, the {\causal} relation in the example can be established because if \textit{``Europe planned for getting stricken Albanian back''} happens, the \textit{``meeting''} will definitely happen in this context.

\paragraph{Conditional:} A pair of events ($e_i, e_j$) exhibits a {\condition} relation \textbf{if $e_i$ facilitates, but may not necessarily leads to $e_j$} according to the given passage. For example, the expectation of \textit{``the dispatch of a multinational force''} is to \textit{``pull Albania back from the brink''}; in other words, the former event can help but does not guarantee the occurrence of the latter one. Therefore, the relation between this pair of events is {\condition}.

\paragraph{Counterfactual:} \textbf{$e_j$ may happen if $e_i$ does not happen}; in other words, if the negation of $e_i$ facilitates $e_j$, then ($e_i, e_j$) has a {\counter} relation. In our example, if \textit{``Europe didn't support Albania,''} which is a negation of what happens in the passage, then \textit{``oust President Sali''} by the \textit{``armed rebels''} would likely happen.

\paragraph{Sub-event:} There is a semantic hierarchy where \textbf{a complex event $e_k$ consists of a set of sub-events $\{e_{k,1}, ..., e_{k,j}, ..., e_{k,n}\}$}. In {\subevent} relations, we require not only $e_{k,j}$'s trigger word to be semantically contained in $e_{k}$'s trigger, but also the arguments of $e_{k,j}$ are either identical or contained in the associated arguments of $e_k$. For example, considering the complex event \textit{``efforts to pull Albania back,''} and its sub-event \textit{``aid is brought into the chaotic Balkan state''}, the trigger ``brought'' is a part of the ``efforts.'' Both subjects are ``Europe,'' both objects / locations are ``Albania'' or ``Balkan state'' and their time can be inferred to be (nearly) identical in the passage. Note that this definition is similar to the event hierarchical structure definition in RED, but stricter than the ``Spatial-temporal containment'' definition in HiEve.
 
\paragraph{Coreference:} \textbf{$e_i$ co-refers to $e_j$ when two events are mutually replaceable}. This requires 1) their event triggers are semantically the same and 2) their event arguments are identical. In our example, the event triggers in the question \textit{``pull'' (back from the brink)} and in the answer \textit{``getting'' (back on to its feet)} are semantically the same. They also share the same subject - Europe, and object - Albania. Their time and location can be inferred from the passage to be the same. Therefore, these two events form a {\coref} relation.  

\section{Related Work}
\label{sec:related}
We briefly survey related work in this section in order to provide broader background over the two key components of {\dataset}: 1) event semantic relations and 2) event-centric reading comprehension.

\subsection{Event Semantic Relations}
Event semantic relations have been studied before and most of them leverage relation extraction formulation for annotations. Causality is one of the widely studied event semantic relations. \citet{mirza-tonelli-2014-analysis, mirza-etal-2014-annotating} follow the \cause, {\enable} and {\prevent} schema proposed by \citet{Wolff-2007} where the first two relations align with our definitions in {\dataset}. \citet{do-etal-2011-minimally} adopted a minimally supervised method and measure event causality based on pointwise mutual information of predicates and arguments, which resulted in denser annotations than previous works.

{\hieve} \cite{glavas-etal-2014-hieve} defines pairwise {\subevent} relation as spatiotemporal containment, which is less rigorous than our definitions where we require containment for all event arguments (subject, object, time and location). Our definition of {\coref} is nearly identical as {\hieve} where two co-referred events denote the same real-world events. \citet{yao-etal-2020-weakly} utilized a weakly-supervised method to extract large scale {\subevent} pairs, but the extracting rules can result in noisy relations.

{\red} \cite{ogorman-etal-2016-richer} proposed to annotate event temporal and semantic relations (\causal, \subevent) jointly. However, due to the complexity of the annotation schema, the data available for semantic relations are relatively sparse. \citet{mostafazadeh-etal-2016-caters} and \citet{caselli-vossen-2017-event} annotate both event temporal and semantic relations in ROCStories \cite{mostafazadeh-etal-2016-corpus} and Event StoryLine Corpus \cite{caselli-vossen-2017-event} respectively. {\dataset} differs from these works by disentangling temporal from other semantic relations and focusing on MRC to capture five proposed event semantic relations.

\subsection{Event-centric MRC}
Datasets leveraging natural language queries for event-centric machine reading comprehension have been proposed recently \cite{zhou-etal-2019-going, ning-etal-2020-torque}. However, they focus on event temporal commonsense, whereas {\dataset} studies other event semantic relations. \citet{du-cardie-2020-event} and \citet{liu-etal-2020-event} reformulate event extraction data as QA tasks to detect event triggers and arguments in a short passage. However, they did not propose new data, and knowing event triggers and arguments are merely a sub-task in {\dataset}, which require both event detection and relation understanding. 

\section{Data Collection}
\label{sec:data}
In this section, we show our data collection procedure and describe the details of our approach to control annotation quality, including qualification exams and steps to validate and train workers.

\subsection{Passage Preparation}
\label{sec:passage-prep}
Passages are selected from news articles in TempEval3 (TE3) workshop \citep{uzzaman-etal-2013-semeval} with initial event triggers provided. We extracted 3-4 continuous sentences that contain at least 7 event triggers. Our choice of the number of sentences is based on previous studies that hierarchical relations such as {\subevent} and {\coref} are likely to span over multiple sentences, but the majority of them are contained within 3-4 sentences \citep{glavas-etal-2014-hieve, ogorman-etal-2016-richer}.
 
\subsection{Main Procedure}
 We use Figure~\ref{fig:mrc-exp} to illustrate our main data collection procedure, which consists of two components: event selection and QA annotations. The actual interface can be found in the appendix.

\vspace{3pt}
\mypar{Event Selections.} Annotators are presented with a passage and initial event trigger annotations. They are allowed to modify event trigger selections per our definition in Section~\ref{sec:define} by highlighting words. These correspond to the highlighted words in the passage of Figure~\ref{fig:mrc-exp}. Our focus is not event extraction, and thus we do not require workers to identify all triggers as some of them are not used in their QAs. Rather, the event selection serves as a warm-up step for the following QA annotations by 1) helping workers locate where desirable events are and 2) ensuring that all the annotated question-answer pairs include events in the passage so that their QAs reason about event relations.
%2) ensuring all questions and answers include an event in the passage so that the QAs are reasoning about events.

\vspace{3pt}
\mypar{QA Annotations.} As the five questions in Figure~\ref{fig:mrc-exp} show, users must ask natural language questions that contain a highlighted event trigger. In order to make questions natural, we allow workers to use different textual forms of an event trigger in the questions, such as ``teach'' v.s. ``taught'' and ``meeting'' v.s. ``meet.'' After writing a question, users need to pick the event semantic type (the blue boxes in Figure~\ref{fig:mrc-exp}) that they reason about, and then select the corresponding answer spans from the passage. If there are multiple answers, we instruct users to select \textit{all} of them. All answers must include an exact highlighted event trigger, and we prohibit answers with more than 12 words to ensure conciseness. We pay \$7.5 for an assignment where annotators need to ask at least five questions using two passages. 
% The only semantic type that is not required to contain a highlighted trigger is \subevent, where we allow annotators to ask a question using an imaginary but concrete event trigger word. This exception enables easier composition of {\subevent} QAs as we found in our pilot studies that complex events do not always exist in a given passage despite the abundance of sub-events.

% \ihung{I think the last sentence is not so informative and if we need more space, we can cut it.}

\begin{figure}[t]
    \centering
\includegraphics[trim=2.5cm 7.5cm 2.5cm 7.5cm, clip, width=0.8\columnwidth]{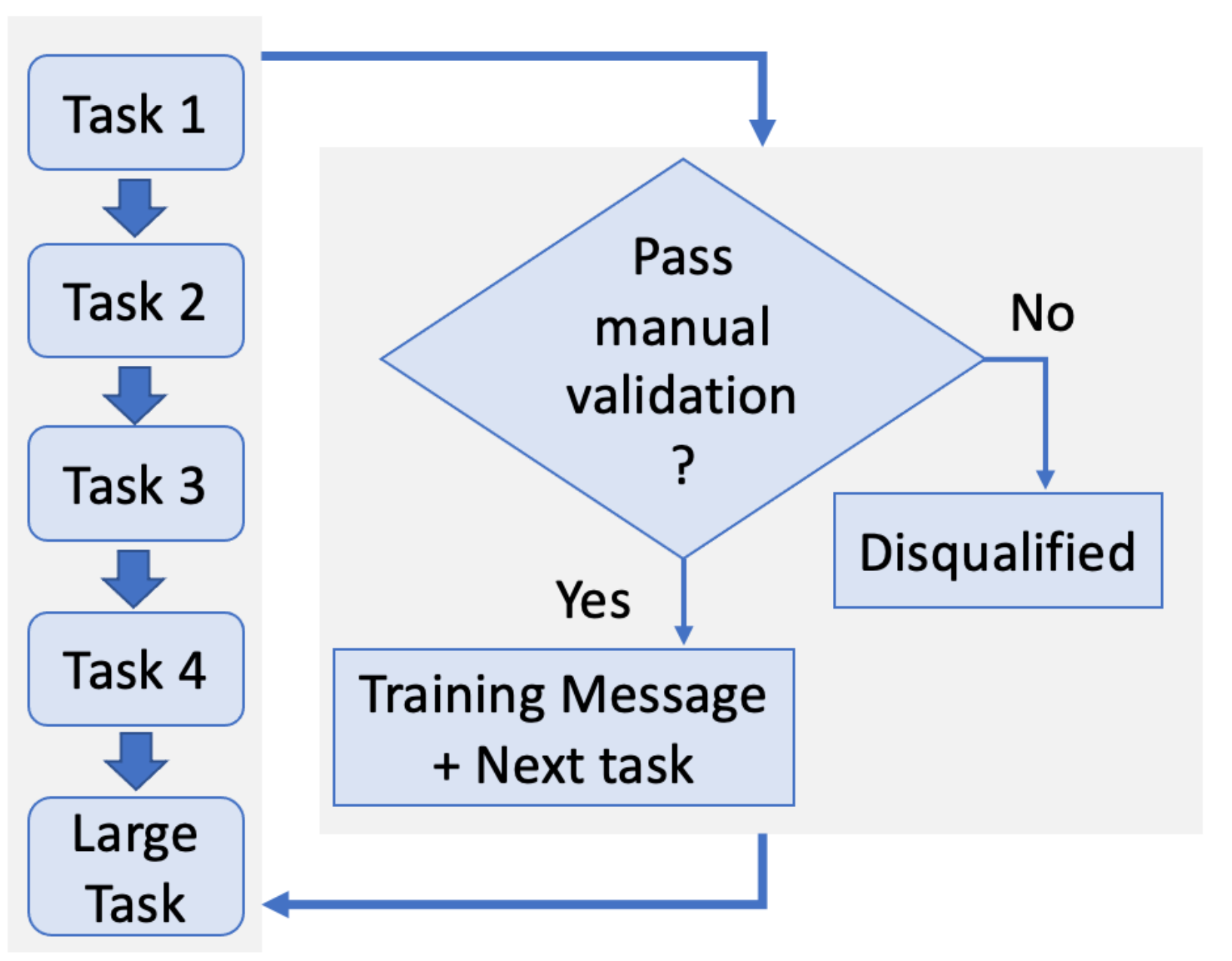}
\caption{An illustration of our quality control, worker validation and training process.}
\vspace{-0.5cm}
\label{fig:valdation}
\end{figure}

\subsection{Quality Control}
\mypar{Qualification.} The initial worker qualification was conducted via an examination in the format of multiple-choice questions hosted by CROWDAQ platform \citep{ning-etal-2020-easy}. We created a set of questions where a passage and a pair of QA are provided, and workers need to judge the correct type of this QA from six choices, including the five defined event semantic relations, plus an invalid option\footnote{A full list of QA validity can be found in Appendix B.}. This examination intends to test workers' skills to 1) distinguish valid QAs from invalid ones based on our definitions; 2) judge the differences for the five proposed \textbf{event semantic relations}.
%\IHNote{we never mention what's ``invalid'' in the paper before. A bit worry about this.}

We recruit workers via Amazon Mechanical Turk with basic qualifications including: 1) at least 1K HITs\footnote{HIT is an assignment unit on Amazon Mechanical Turk.} approved; 2) at least 97\% approval rate. A single qualification exam consists of 10 multiple-choice questions. Participants are given 3 attempts to pass with a $>=0.6$ score. We found this qualification examination effectively reduces the rate of spammers to nearly 0\%.

\vspace{3pt}
\mypar{Worker Validation and Training.} Since the real task is much more challenging than the qualification exams, we adopted a meticulous five-stage worker validation and training process to ensure data quality. As Figure~\ref{fig:valdation} shows, for workers who passed the qualification exams, we repeat the validation and training steps four times %illustrated in Figure~\ref{fig:valdation} 
until workers reach the final large tasks.

% \violets{do we have quantitative measurement for ``showed a serious misunderstanding''?}

In each validation and training step, two of our co-authors independently judge workers' annotations to determine 1) whether a provided QA pair is valid per our definitions and 2) whether the answers provided are complete. Typically, we disqualify workers whose QA validity rate falls below 90\%. Exceptions are given upon careful examination and reviewer discussion. For workers who pass a manual validation, we write a training message correcting all errors they made and invite them to the next task. We also add missing answers as a part of the validation process and reserved the validated annotations as our evaluation data.

There are 1, 2, 3, 10, and 25 HITs in Task 1-4 and Large Task respectively. For Task 1-3, we validate all QAs, and for Task 4, we randomly select 20\% questions per worker to validate. In order to work on the final large task, a worker needs to maintain an average QA validity rate higher than 90\%. We further request one co-author to validate all questions with passages overlapped with the validated data above. This ensures that there are no passage overlaps between the training and evaluation data. All author validated data comprise our final evaluation data in the experiments.
%\ihung{we probably need a footnote to define what is a HIT here to avoid the confusion between this one and the HIT@1.}

%\violets{what does ie mean by ``in the following section'' here?}

% \violets{what does this mean? what's the purpose of doing this step?}

\section{Data Analysis}
\label{sec:statistics}
 Our passage preparation (Section~\ref{sec:passage-prep}) produces 4.3K passages in total with 1887 of them randomly selected and annotated. We collect 6018 questions from 70 workers using Amazon Mechanical Turk and 1471 of them fully validated by co-authors as the evaluation set. We further split our evaluation data into dev and test sets based on passages. The remaining data are used as the training set. A summary of data statistics is shown in Table~\ref{tab:data}. 
%\IHNote{Maybe also good to know how many different paragraph?.}

\begin{table}[htbp!]
\centering

\scalebox{0.7}{
\setlength{\tabcolsep}{5pt}
\begin{tabular}{l|ccc}
\toprule
& Train & Dev & Test \\
\midrule
\# of Passages & 1492 & 108 & 287 \\
\midrule
\# of Questions - Overall & 4547 & 301  & 1170 \\
\midrule
- {\causal} & 2047 & 118 & 431\\
- {\condition} & 928 & 58 & 289 \\
- {\counter} & 294 & 28 & 106 \\
- {\subevent} & 678 & 59 & 204\\
- {\coref} & 600 & 38 & 140 \\
\bottomrule
\end{tabular}
}
\caption{Passages and questions (overall + type breakdown) statistics for different data splits.}
\label{tab:data}
\vspace{-0.5cm}
\end{table}

\subsection{Type Distribution}
\begin{figure}[t]
    \centering
\includegraphics[trim=0cm 0cm 0cm 0cm, clip, width=0.9\columnwidth]{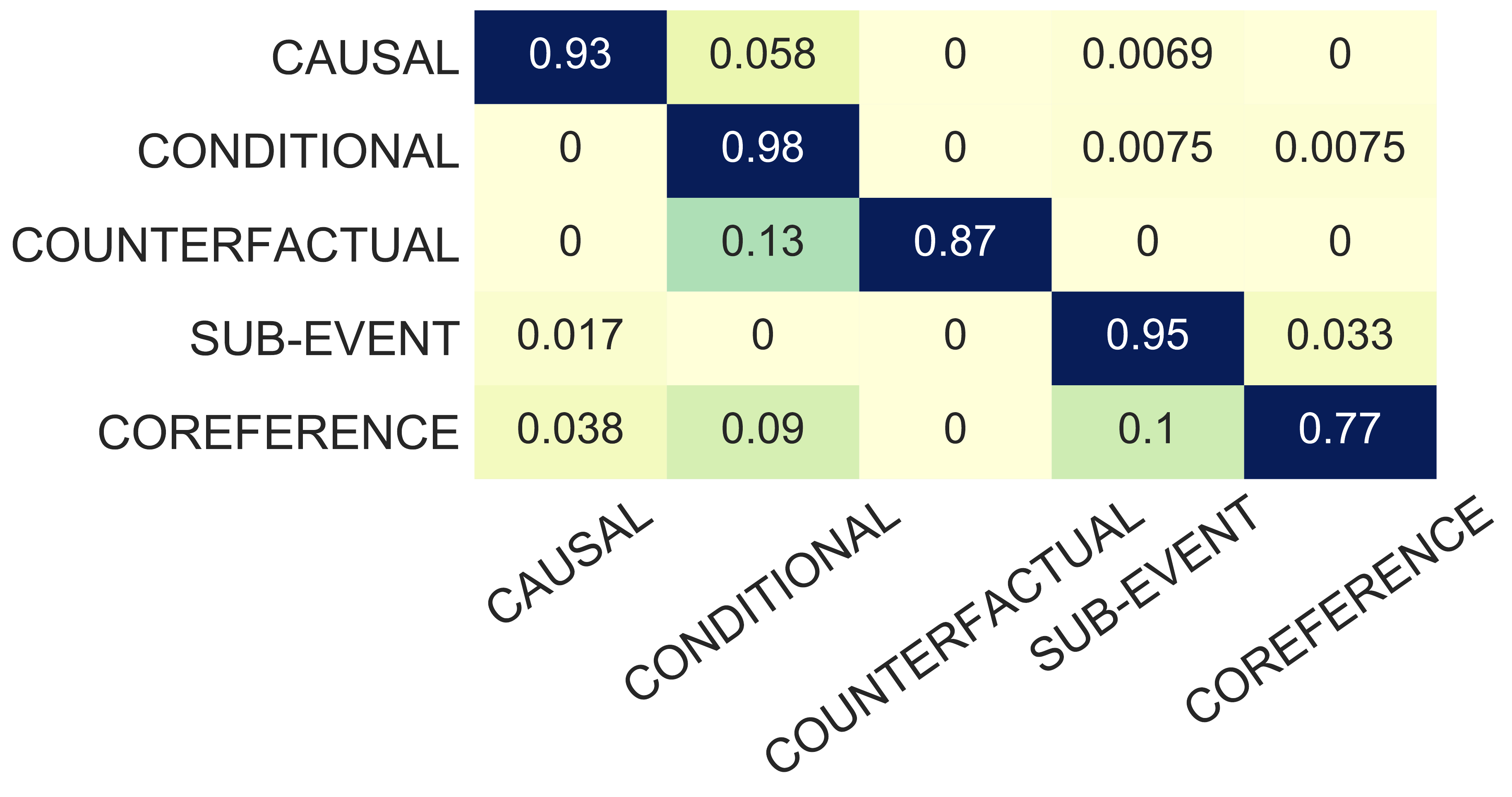}
\vspace{-.5cm}
\caption{Type confusion matrix between workers' original annotations  and the majority votes after co-authors' validation. Rows are annotators' types whereas columns are the majority votes.}
\label{fig:type-confusion}
\vspace{-0.5cm}
\end{figure}

As we can observe in Table~\ref{tab:data} and Figure~\ref{fig:type-dist-train} -~\ref{fig:type-dist-dev} in the appendix, {\dataset} consists of 64.2\% {\causal} and {\condition} questions. In Figure~\ref{fig:type-confusion}, we further show the type disagreements using data validated by two co-authors. The rows indicate workers' original types and the columns are the majority votes between the annotators and co-authors.
%\IHNote{might be confusing since we said 2 co-authors. So, probably need to clarify the majority is including the turker's annotation?}
%\IHNote{Maybe highlight co-author's definition is more tight?}
%\IHNote{not sure it's easy for general audience to understand double negation}
As we can observe, the matrix is dominated by diagonal entries. Some noticeable disagreements are 1) between {\causal} and {\condition} where people have different opinions on the degree of causality between events; 2) between {\counter} and {\condition} as some {\counter} questions, with double negations\footnote{Double negated questions have the form of \textit{``what will not happen if Event A does not happen''}}, are merely {\condition};  3) between {\coref} and {\subevent} where annotated co-referred events do not have identical event arguments according to co-authors' judgements. These results align with previous studies that some event semantic relations are inherently hard to distinguish \cite{glavas-etal-2014-hieve, ogorman-etal-2016-richer}. 

\vspace{3pt}
\mypar{Type Agreements.} The inter-annotator-agreement (IAA) score is 85.71\% when calculated using pair-wise micro F1 scores, and is 0.794 per Fleiss's $\kappa$\footnote{0.794 implies substantial agreement\cite{10.2307/2529310}. The detailed calculation can be found in appendix C.}. The IAA scores are calculated using the same data reported in Figure~\ref{fig:type-confusion}. The high IAA scores demonstrate strong alignments between annotators and co-authors in judging event semantic relations.

\subsection{Other Statistics}
We show n-grams in questions and the number of answers below. More analysis on tokens and worker distributions can be found in Appendix-\ref{sec:worker-dist}~\ref{sec:num-tokens}.

\begin{figure}[h!]
    \centering
    \begin{subfigure}[b]{0.7\columnwidth}
        \centering
        \includegraphics[width=\columnwidth]{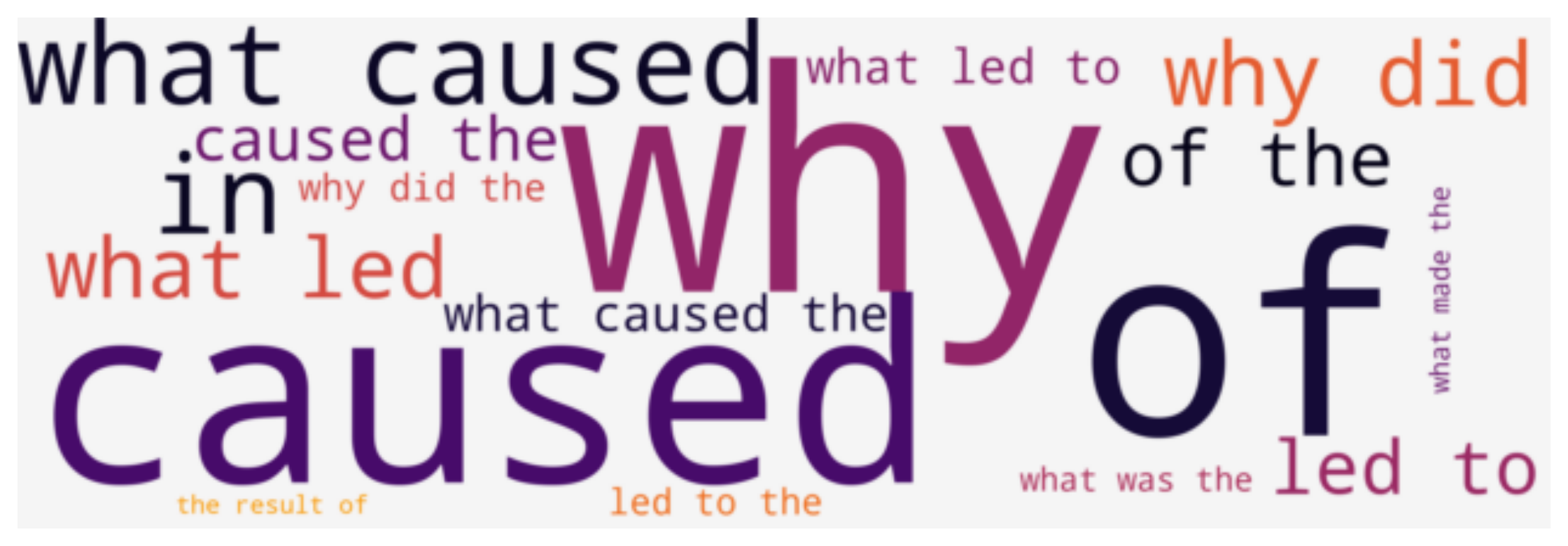}
        %\caption{Most frequent n-grams: for {\causal}}
        %\label{fig:most_freq_causal} 
    \end{subfigure}
        \begin{subfigure}[b]{0.49\columnwidth}
        \centering
        \includegraphics[width=\columnwidth]{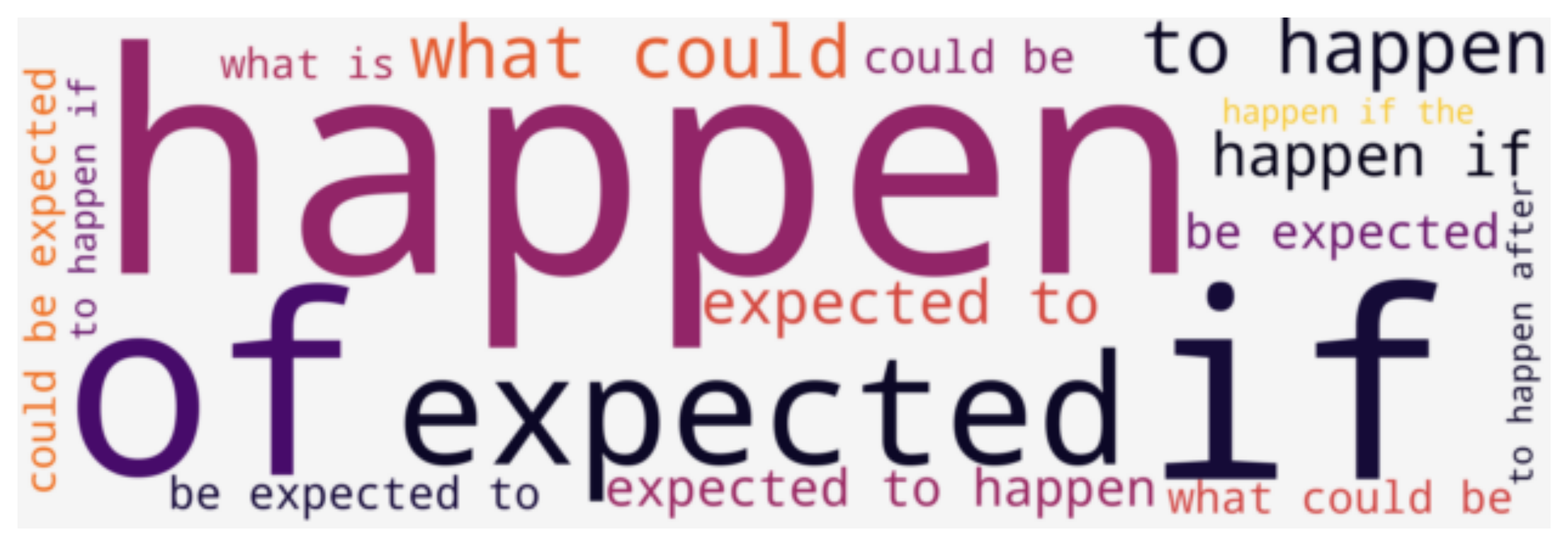}
        %\caption{Most frequent n-grams: for {\causal}}
        %\label{fig:most_freq_causal} 
    \end{subfigure}
        \begin{subfigure}[b]{0.49\columnwidth}
        \centering
        \includegraphics[width=\columnwidth]{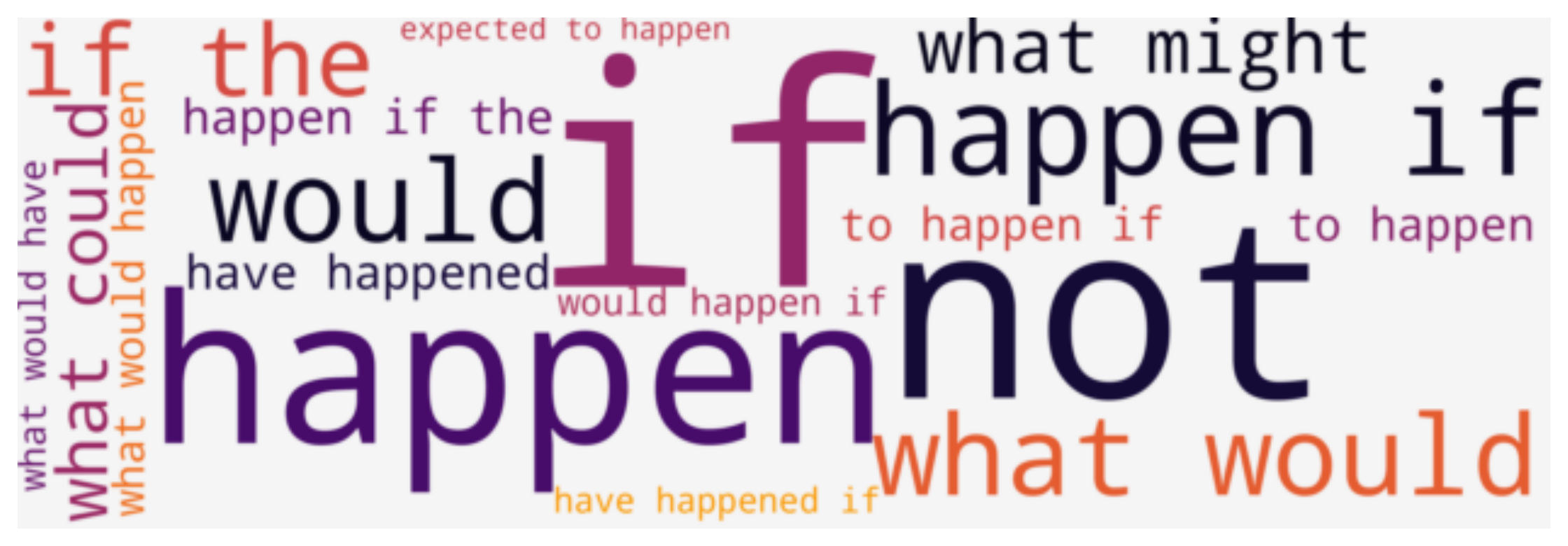}
        %\caption{Most frequent n-grams: for {\causal}}
        %\label{fig:most_freq_causal} 
    \end{subfigure}
        \begin{subfigure}[b]{0.49\columnwidth}
        \centering
        \includegraphics[width=\columnwidth]{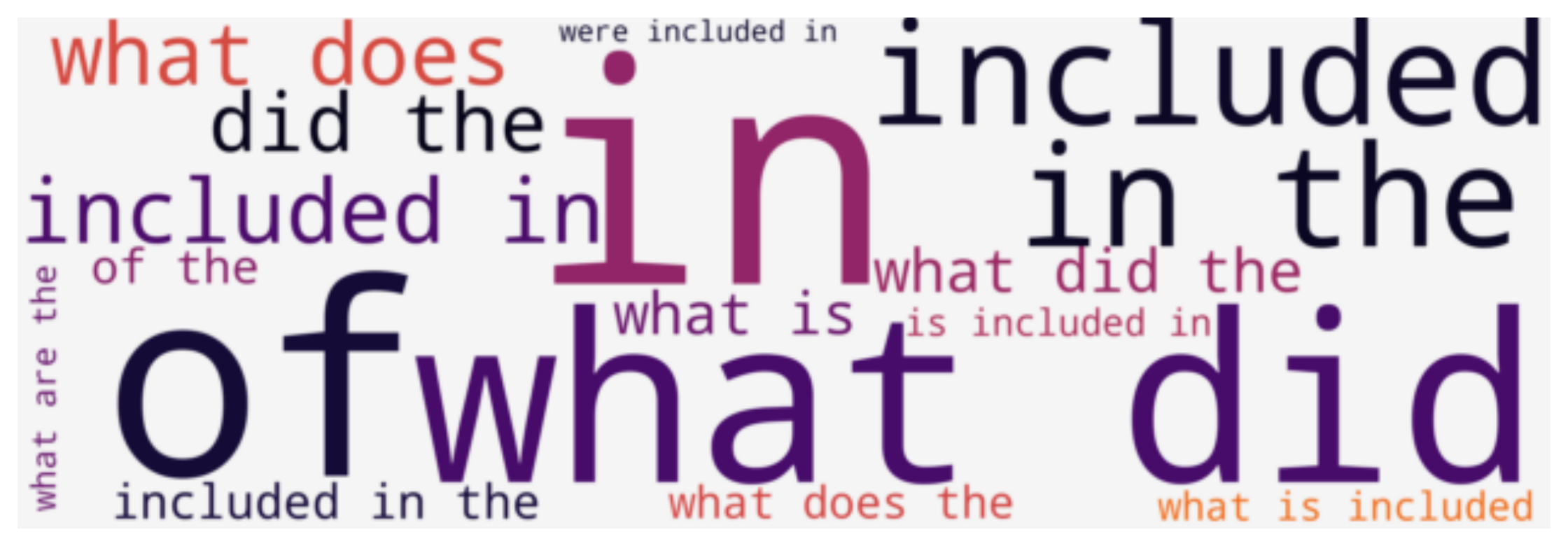}
        %\caption{Most frequent n-grams: for {\causal}}
        %\label{fig:most_freq_causal} 
    \end{subfigure}
        \begin{subfigure}[b]{0.49\columnwidth}
        \centering
        \includegraphics[width=\columnwidth]{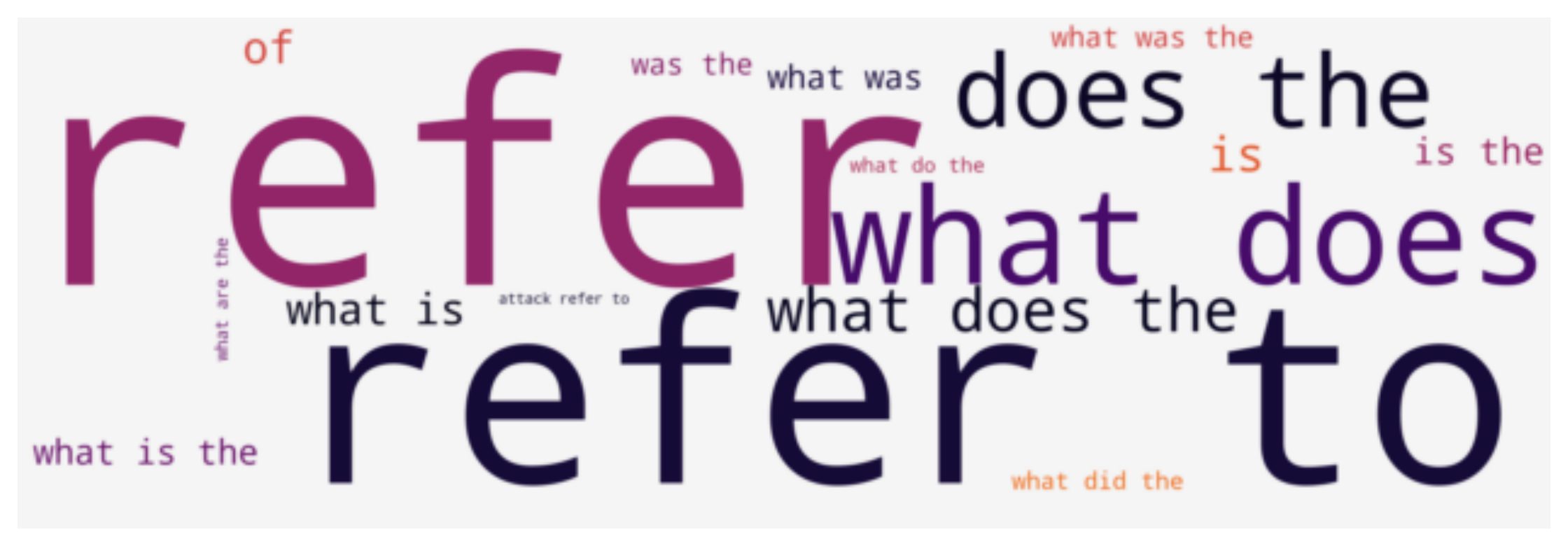}
    \end{subfigure}
    \vspace{-0.8cm}
    \caption{Most frequent n-grams in questions for each semantic type. First row: {\causal}; second row: {\condition} + {\counter}; third row: {\subevent} + {\coref}.}
    \vspace{-0.3cm}
    \label{fig:most_freq_all}
\end{figure}

% \violets{I think is this super interesting. Is it possible that we put all 5 types in a span column figure (each one can be smaller). Also, can we remove "stop words" i.e., the ones that have high frequency in all typs. e.g., what.}

\mypar{Frequent N-grams in Questions.} Figure~\ref{fig:most_freq_all} illustrates the most frequent unigram, bigram and trigrams in each type of questions after removing non-informative stop-words. These n-grams can be considered as semantic cues in the questions to reason about particular semantic relations. For example, `why' and `what caused' imply strong causality; `included' indicates containment of events; `not' in {\counter} indicates negation of events.

\vspace{3pt}
\mypar{Number of Answers.} Table~\ref{tab:ans-num} shows the average number of answers for each semantic type. {\subevent} contains the most answers, which aligns with our intuition that a complex event in the passage often contains multiple sub-events. The evaluation sets contain about 0.5 answers more than the training set as co-authors added the missing answers in the validation process. Considering each unique question and answer as an event, {\dataset} captures 10.1K event pairs, which are larger than previous RE datasets such as RED and HiEve.

\begin{table}[htbp!]
\centering
\scalebox{0.7}{
\setlength{\tabcolsep}{5pt}
\begin{tabular}{l|ccc}
\toprule
Semantic Types & Train & Dev & Test  \\
\midrule
{\causal} & 1.3 & 1.5 & 1.9\\
{\condition} &  1.3 & 1.9 & 2.0\\
{\counter} & 1.2 & 1.3 & 1.7 \\
{\subevent} & 3.0 & 3.6 & 3.1 \\
{\coref} &  1.2 & 1.2 & 1.6 \\
\bottomrule
\end{tabular}
}
\vspace{-0.2cm}
\caption{Average number of answers by semantic types.}
\label{tab:ans-num}
\vspace{-0.25cm}
\end{table}

\begin{table*}[htbp!]
\centering
\scalebox{0.83}{
\setlength{\tabcolsep}{4.3pt}
\begin{tabular}{l|ccc|ccc}
\toprule
 & \multicolumn{3}{c|}{\textbf{Dev}} & \multicolumn{3}{c}{\textbf{Test}} \\
\cmidrule(lr){2-4}\cmidrule(lr){5-7}
 & $F_1^T$ & \textbf{HIT@1} & \textbf{EM} & $F_1^T$ & \textbf{HIT@1} & \textbf{EM}\\
\midrule
Generative Zero-shot: T5-base & 18.0 & 55.8 & 0.0 & 21.1 & 61.0 & 0.0 \\
Generative Zero-shot: UnifiedQA-base & 49.0 & 61.5 & 10.6 & 46.5 & 61.5 & 7.1 \\
Generative Zero-shot: UnifiedQA-large & 51.1 & 69.4 & 14.3 & 48.7 & 66.5 & 9.7\\
\midrule
Generative Fine-tune: BART-base  & 53.1($\pm$0.4) & 66.9($\pm$1.7) & 14.1($\pm$1.0) & 53.3($\pm$0.8) & 68.1($\pm$1.2) & 15.1($\pm$0.7)\\
Generative Fine-tune: BART-large & 57.2($\pm$1.0) & 72.1($\pm$1.4) & 15.1($\pm$2.1) & 56.1($\pm$1.0) & 71.5($\pm$2.2) & 15.2($\pm$0.9)\\
Generative Fine-tune: T5-base & 63.2($\pm$1.1) & 80.8($\pm$1.7) & 22.1($\pm$0.9) & 58.5($\pm$0.7) & 76.2($\pm$1.0) & 20.5($\pm$0.9) \\
Generative Fine-tune: UnifiedQA-base & 64.6($\pm$0.4) & 82.0($\pm$0.4) & 23.8($\pm$1.0) & 59.3($\pm$0.2) & 78.1($\pm$0.4) & 20.6($\pm$0.5) \\
Generative Fine-tune: UnifiedQA-large & 66.8($\pm$0.2) & \textbf{87.2($\pm$0.3)} & \textbf{24.4($\pm$0.3)} & 63.3($\pm$0.8) & \textbf{83.5($\pm$0.7)} & \textbf{22.1($\pm$0.4)}\\
\midrule
Extractive Fine-tune: RoBERTa-large & \textbf{68.8($\pm$0.7)} & 66.7($\pm$1.1) & 16.7($\pm$0.2) & \textbf{66.1($\pm$0.2)} & 63.8($\pm$1.6) & 15.9($\pm$0.5) \\
\midrule
Human Baseline & - & - & - & 79.6 & 100 & 36.0 \\
\bottomrule
\end{tabular}
}
\vspace{-0.2cm}
\caption{Experimental results for answer generation. All numbers are 3-seed average with standard deviation reported, except for human baseline and zero-shot performances. All models refer to the generative QA task except for RoBERTa-large, which we use for the extractive QA task. Statistical tests are shown in Appenidx~\ref{sec:significance-test}.} 
\label{tab:results-answer-gen}
\vspace{-0.5cm}
\end{table*}

\section{Experimental setup} 
% \violets{add some discussion here to explain the goal of the experiments. e.g., benchmark the performances of current models on this dataset, understand the challenge, and point to possible future research directions for semantic relation understanding.}
\label{sec:experiments}
We design experiments to provide benchmark performances and understand learning challenges to facilitate future research on {\dataset}. We formulate our QA task as a conditional answer generation problem. This choice is inspired by recent works such as UnifiedQA \citep{khashabi-etal-2020-unifiedqa} that achieve impressive outcomes by integrating various QA tasks (extractive, abstractive and multiple-choice) as a single generative QA pre-training task. \citet{li2021document} and \citet{paolini2021structured} also show that by reformulating original extractive tasks as generation tasks, it enables models to better exploit semantic relations between context and labels as well as the dependencies between different outputs. To better demonstrate the benefits of the proposed generative QA task, we compare it with a traditional extractive QA task.
We introduce our experimental design and evaluation metrics subsequently.

\subsection{Generative QA}
Given a question $q_i$ and a passage $P_i = \{x_1, x_2, ... x_j, ... x_n\}$ where $x_j$ represents a token in the passage, the answer generation task requires the model to generate natural language answers $A'_i = \{a'_{i,1}...a'_{i,k}\}$. For the gold answers $A_i = \{a_{i,1}...a_{i,k}\}$, each answer span $a_{i,k} \in P_i$. We follow the input format of UnifiedQA \citep{khashabi-etal-2020-unifiedqa} by concatenating $q_i$ and $P_i$ with a ``\textbackslash n'' token. For training labels, we concatenate multiple answers with a ``;'' token.

\subsection{Extractive QA}
Given $q_i$ and $P_i$, this task requires a model to predict whether each token $x_j \in P_i$ is an answer or not. Following the ``B-I-O'' labeling conventions in the IE field, we create a vector of labels with `2' if $x_j$ is the beginning token of an answer span;  `1' if $x_j$ is an internal token of an answer span; `0' if $x_j \notin A_i$. The input is the same as generative QA except that we concatenate $q_i$ and $P_i$ with two ``<$\backslash s$>'' tokens to be consistent with the pair-sentence input format of the base model, RoBERTa-large \cite{ROBERTA-19}.

To compare fairly with the generative QA task, we construct candidate answer spans by examining predicted labels for all tokens. Both ``BI*'' and ``I*'' cases are considered as valid answers. Finally, we map positive answer tokens' ids back to natural language phrases. More formally, we can denote the final candidate answers of the task as $A''_i = \{a''_{i,1}...a''_{i,k}\}$, where $a''_{i,k} \in P_i$.

\subsection{Evaluation Metrics.} It is important to assess how well models can find all valid answer. We evaluate this by using token-based $F_1$ and exact-match measures. On the other hand, when interacting with machines, we would like the top answer returned to be correct. We measure this by \textbf{HIT@1} scores.
\begin{itemize}
    \item Let $U_i, U'_i$ denotes all uni-grams in $A_i, A'_i$. We have $F_1^{T} = \frac{2*P*R}{P + R}$ where $P = \frac{|U_i \cap U'_i|}{|U'_i|}, \quad  R = \frac{|U_i \cap U'_i|}{|U_i|}$.
    \item \textbf{HIT@1} equals to 1 if the top predicted answer, i.e. $a'_{i,1}$ or $a''_{i,1}$ contains a correct event trigger; otherwise it is 0. This metrics is well defined as all questions in our data contain at least an answer and all (well trained) models return at least one answers. For both generative and extractive QAs, we use the leftmost answer as the top answer. 
    \item \textbf{EM} or exact-match equals to 1 if $\forall a'_i \in A'_i, a'_i \in A_i$ \textbf{and} $\forall a_i \in A_i, a_i \in A'_i$; otherwise, \textbf{EM} = 0.
\end{itemize}

\subsection{Baselines}
\mypar{Model Baselines.} For our primary generation QA task, we fine-tuned several sequence-to-sequence pre-trained language models on {\dataset}: BART \citep{lewis-etal-2020-bart}, T5 \citep{2020t5} and UnifiedQA. As mentioned, UnifiedQA (based on BART and T5) is pre-trained on various QA tasks. It also demonstrates powerful zero-shot learning capabilities on unknown QA tasks, which we tested on {\dataset} too. Due to computation constraints, the largest model we are able to finetune is UnifiedQA (T5-large). We leave further investigation to future modeling studies. 

Since extractive QA can be considered as a token prediction task, we build our model based on RoBERTa-large with token mask prediction pre-training objectives. Models and fine-tuning details can be found in Appendix~\ref{sec:model-details}.

\vspace{3pt}
\mypar{Human Baselines.} To show the human performance on the task, we randomly select 20 questions for each semantic type from the test set. Two co-authors provide answers for these questions, and we compare their mutually agreed answers with the original answers. We ensure co-authors never saw these questions previously. $F_1^T$, \textbf{HIT@1} and \textbf{EM} scores are calculated as the human performances.
\section{Results and Analysis}
\label{sec:results}
In this section, we present and analyze results for the experiments described in Section~\ref{sec:experiments}.

\subsection{Generative QA}
As Table~\ref{tab:results-answer-gen} shows, UnifiedQA-large achieves the best average performances among all generative QA baselines, with 63.3\%, 83.5\% and 22.5\% for $F_1^T$, \textbf{HIT@1} and \textbf{EM} scores on the test set, which are 16.3\%, 16.5\% and 13.1\% below the human performances. We also observe that UnifiedQA-base with 220M parameters outperforms other comparable or larger models such as T5-base and BART-large with 2-3x more parameters, showing the effectiveness of pre-training with generative QA tasks.

\begin{figure}[t]
    \centering
\includegraphics[trim=0cm 0cm 0cm 0cm, clip, width=0.98\columnwidth]{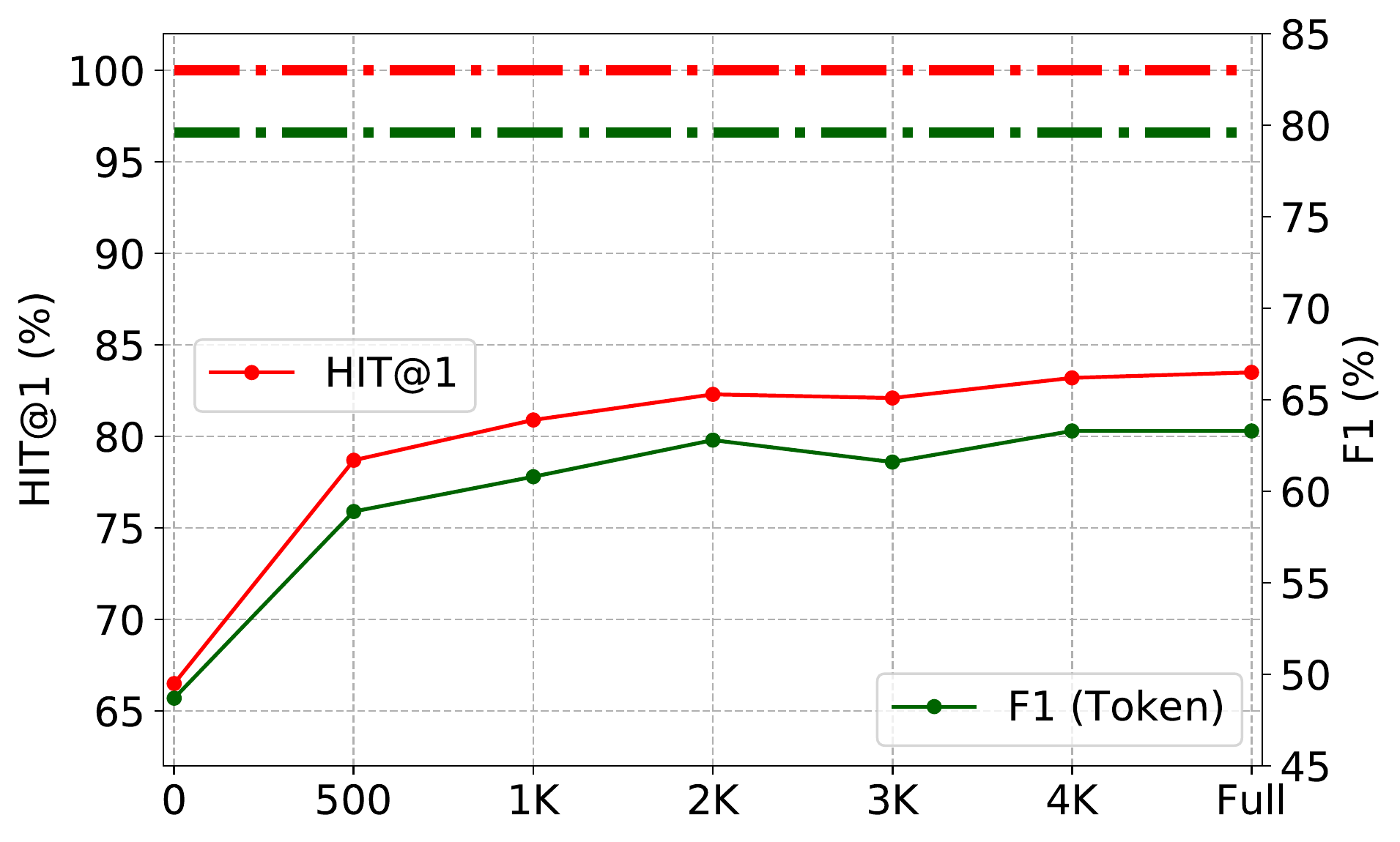}
\vspace{-0.3cm}
\caption{Fine-tuning UnifiedQA-large results by using 500, 1K, 2K, 3K, 4K and full train data. Dashed lines on the top are corresponding human performances.}
\label{fig:downsample}
\vspace{-0.3cm}
\end{figure}

\vspace{3pt}
\mypar{Zero-shot and few-shot Learning.} UnifiedQA also demonstrates powerful zero-shot and few-shot learning capabilities in a variety of QA tasks. We observe similar patterns where zero-shot learning from UnifiedQA can significantly outperform its T5 counterpart in Table~\ref{tab:results-answer-gen}. For few-shot learning, we show in Figure~\ref{fig:downsample} that fine-tuning with only 500-1K examples, the model can achieve quite comparable results with full-training. The model performances level off as the second half of the training data provide $\le 1.2\%$ improvements across all metrics. This suggests that the benefits of getting more data diminish drastically and data size may not be the bottleneck of learning for \dataset.

\begin{figure}[t]
    \centering
\includegraphics[trim=0cm 0cm 0cm 0cm, clip, width=0.98\columnwidth]{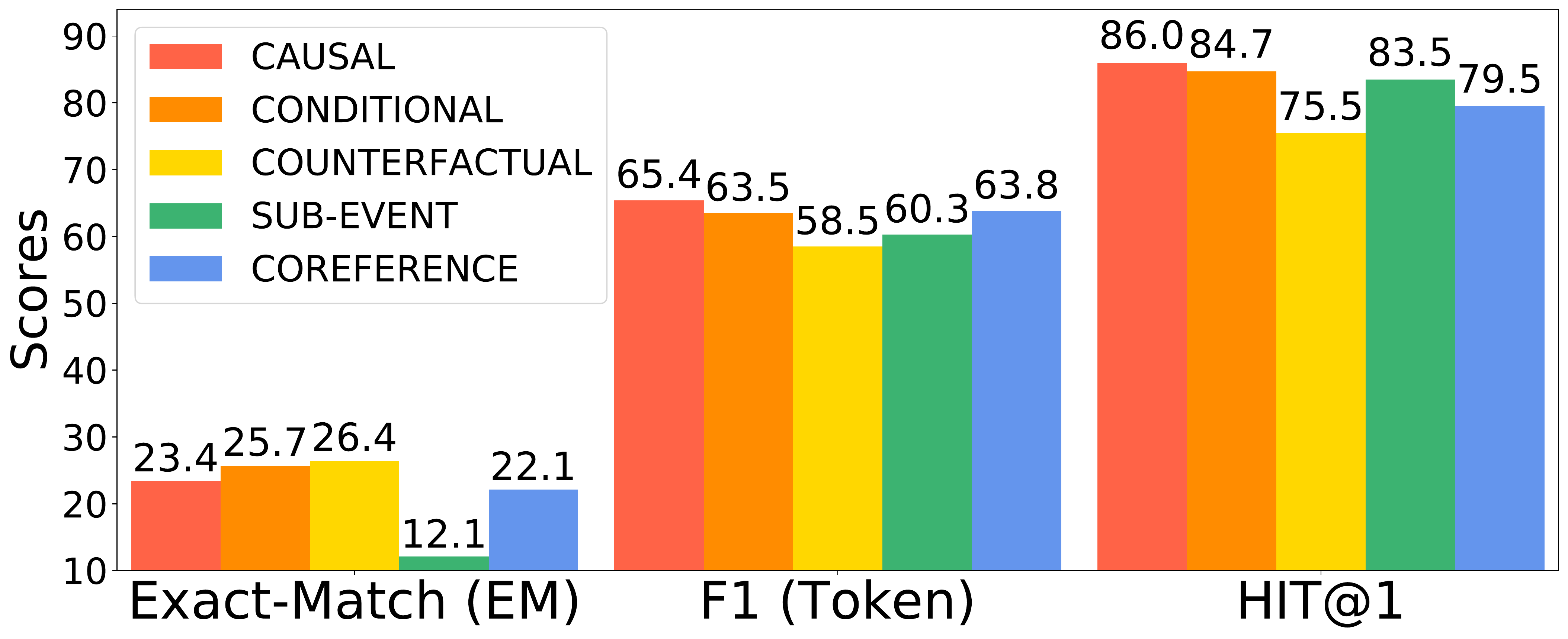}
\vspace{-0.3cm}
\caption{Test performances for each semantic type.}
\label{fig:breakdown_perf}
\vspace{-0.6cm}
\end{figure}

\vspace{3pt}
\mypar{Breakdown performances.} In Figure~\ref{fig:breakdown_perf}, we show performances for each semantic type on the test data. Not surprisingly, {\causal} and {\condition} achieve best performances as they are the more dominant semantic types in {\dataset}. Model training may favor these two types. Interestingly, though {\counter} relation has the smallest number of training questions and requires more complex reasoning than {\condition} due to its negation, our models can learn this relation relatively well per \textbf{EM} and $F_1$ measures. This could be contributed by 1) the similarity between {\counter} and {\condition} relations, and 2) the negations are well detected through textual cues in the model training. On the other hand, the significantly lower \textbf{HIT@1} score for {\counter} suggests that it is challenging for models to pin-point the most confident answer.

Hierarchical relations, {\subevent} and {\coref} in general have lower scores than {\causal} and {\condition}, which could be attributed to two factors: 1) these two categories have smaller percentages (28.1\% combined) in training data; 2) understanding these two relations requires complicated reasoning skills to capture not only the hierarchical relations for event triggers but also for their associated arguments. Figure~\ref{fig:downsample-sub-coref} in the appendix shows the similar plateauing effect of adding more training samples for these two relations, which implies that data size may not be the only factor for weaker performances, and these two semantic relations could be inherently challenging to comprehend.

\paragraph{Answer completeness.} In Table~\ref{tab:ans-num}, we show that the validated data contain about 0.5 more answers per question. Besides some rare obvious misses, \textbf{proximity} and \textbf{saliency} are the two reasons we observe that contribute most to this discrepancy. Our input data include long passages with an average of 128 tokens. Even well-trained workers can overlook relations for event pairs that are physically distant from each other. Moreover, long-distance relations are often less salient. For non-salient relations, expert or external knowledge may be needed to disambiguate. We found workers tend to be conservative by avoiding these non-salient answers.

\begin{table}[htbp!]
\centering
\scalebox{0.75}{
\setlength{\tabcolsep}{4pt}
\begin{tabular}{l|ccccc}
\toprule
& \#Ans. & $F_1^T$ & \textbf{HIT@1} & \textbf{EM}\\
\midrule
original & 1.41 & 58.7($\pm$0.2) & \textbf{78.8($\pm$0.3)} & \textbf{18.8($\pm$0.4)}\\
completed & 1.84 & \textbf{59.3($\pm$0.1)} & 78.5($\pm$0.3) & 16.9($\pm$0.4)\\
\bottomrule
\end{tabular}
}
\vspace{-0.2cm}
\caption{Performances on test data. Workers' original annotations v.s. completed by another worker.}
\label{tab:completed-samples}
\vspace{-0.25cm}
\end{table}

To precisely gauge the impact of answer completeness, we randomly sample 500 questions with type distribution similar to the training data and request qualified workers to find more complete answers. We then retrain UnifiedQA-large with both the original and the more completed answer annotations. Table~\ref{tab:completed-samples} shows that that the ``completed'' set has an average number of answers similar to those in our validated data, but we observe no significant improvements. We hypothesize that 1) through our rigorous validation and training, workers are able to identify important answers; 2) the request to find more complete answers could inadvertently introduce some noise, which cancels out the benefits of increasing answer numbers.

\subsection{Extractive QA}
\label{sec:extractive-qa}
\begin{figure}[t]
    \centering
\includegraphics[trim=0cm 0cm 0cm 0cm, clip, width=0.98\columnwidth]{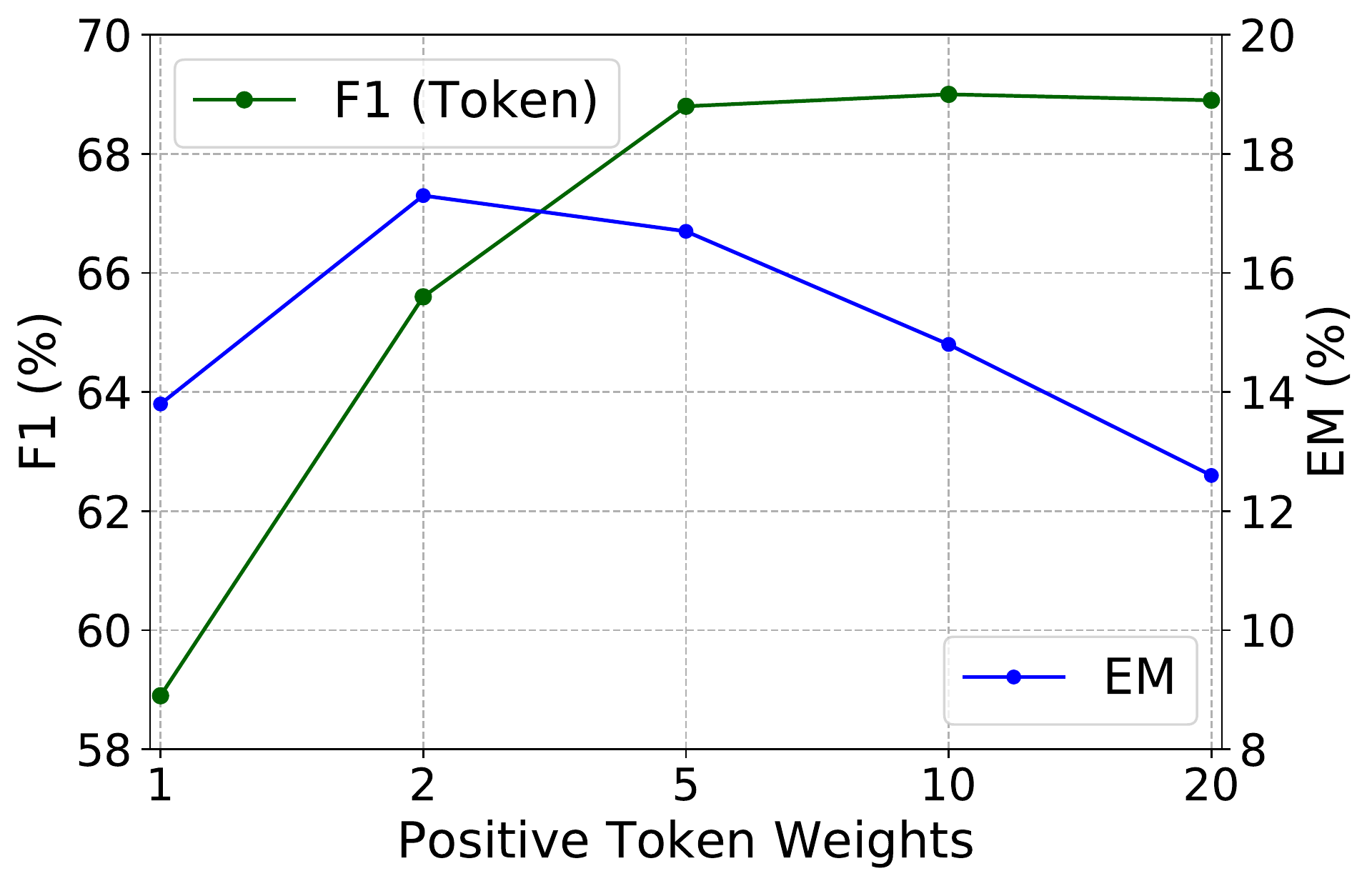}
\vspace{-0.3cm}
\caption{$F_1^T$ v.s. \textbf{EM} scores on the dev set by increasing training weights on positive answer tokens.}
\label{fig:span-detection}
\vspace{-0.3cm}
\end{figure}

In this section, we discuss results for the extractive QA task. In Table~\ref{tab:results-answer-gen}, we observe that extractive QA by finetuning RoBERTa-large achieves the best token F$_1$ scores, yet under-performs generative QA per \textbf{HIT@1} and \textbf{EM} metrics. We further compare $F_1^T$ with \textbf{EM} scores by increasing training weights on positive tokens, i.e. `B' or `I'. Figure~\ref{fig:span-detection} shows that as we train models to focus more on the positive answer tokens, $F_1^T$ keeps increasing up to weight = 10, but answer \textbf{EM} starts to fall after weight = 2. These results imply that extractive QA excels at finding tokens or phrases that resemble or partially overlap with true answers (good $F_1^T$ scores), but falls short on producing complete and meaningful texts that truly represent event spans.

To verify our hypothesis above, we examine real predictions where both the best generative and extractive QA models do not predict exact answers (i.e. per-sample \textbf{EM} = 0). We list several of them in Table~\ref{tab:ex-gen-extract} of the appendix. In general, extractive QA predicts many single or disconnected tokens that are not meaningful, whereas generative QA, despite making wrong predictions, produces answer spans that are complete and coherent. 

To summarize, the comparative studies between generative and extractive QAs emphasize the importance of using multiple metrics to evaluate models and highlight the contribution of leveraging answer generation to solve {\dataset} where complete and meaningful event spans rather than partial tokens are crucial to answer questions.

\subsection{Discussion and Future Research}

\paragraph{Statistical tests.} Modeling is not the main focus of this paper, but we conduct McNemar's tests \citep{McNemar-1947} for comparable models in Table~\ref{tab:significance} of Appendix~\ref{sec:significance-test}. Most of the pairwise tests show strong statistical significance.

\paragraph{Future research.} {\dataset} facilitates a promising research direction of few-shot learning for event semantic relations as a generative QA task, yet remains challenging since large SOTA systems significantly under-perform human baselines. Future research can explore building question generation systems to automatically annotate a larger scale of data or study the possibilities of transfer learning between this MRC data and other event-centric reasoning tasks.

% However, when presented with enough examples comprising of both complete and partial answers, human beings can learn, generalize and thus find complete answers for unseen questions. Learning from partial annotations has been widely explored in several tasks such as named-entity recognition \citep{huang2019learning, shang2018learning}, word segmentation, and part-of-speech tagging \citep{tsuboi-etal-2008-training, yang-vozila-2014-semi}. Therefore, an encouraging future research direction of creatively using our data would be either leveraging or advancing partial signal learning methods in order to close the gap between SOTA models and human performances as shown in Table~\ref{tab:results-answer-gen}.
%\vspace{-0.2cm}
\section{Conclusion}
\label{sec:conclude}
We propose \dataset, an MRC datasets for comprehensive event semantic reasoning. We adopt meticulous data quality control to ensure annotation accuracy. {\dataset} enables a generative question answering task, which can be more challenging than the traditional event relation extraction work. The difficulty of the proposed data and task is also manifested by the significant gap between machine and human performances. We thus believe that {\dataset} would be a novel and challenging dataset that empowers future event-centric research.
\section*{Acknowledgments}
This work is supported by the Intelligence Advanced Research Projects Activity (IARPA), via Contract No. 2019-19051600007 and DARPA under agreement
number FA8750-19-2-0500.

\bibliographystyle{acl_natbib.bst}
\bibliography{anthology.bib, emnlp2021.bib}
\clearpage
\appendix
\section*{Appendix}
\label{sec:appendix}

\section{Interface}
\begin{figure*}[h!]
\centering
    \begin{subfigure}[b]{0.9\textwidth}
    \centering
    \includegraphics[width=\textwidth]{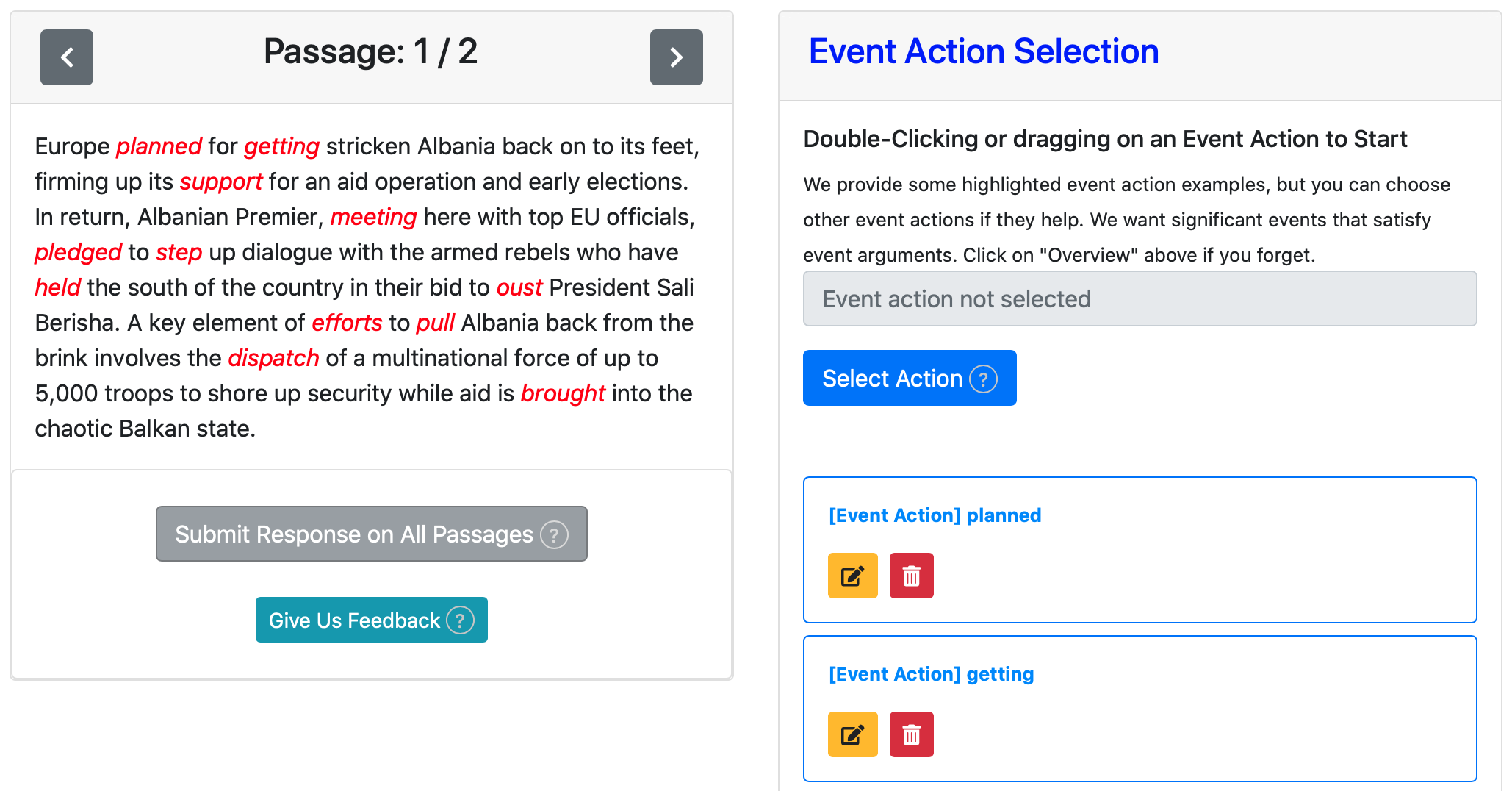}
    \caption{Event Selections in Progress}
    \label{fig:interface-event1}
    \end{subfigure}
    \par\bigskip
    \begin{subfigure}[b]{0.9\textwidth}
    \centering
    \includegraphics[width=\textwidth]{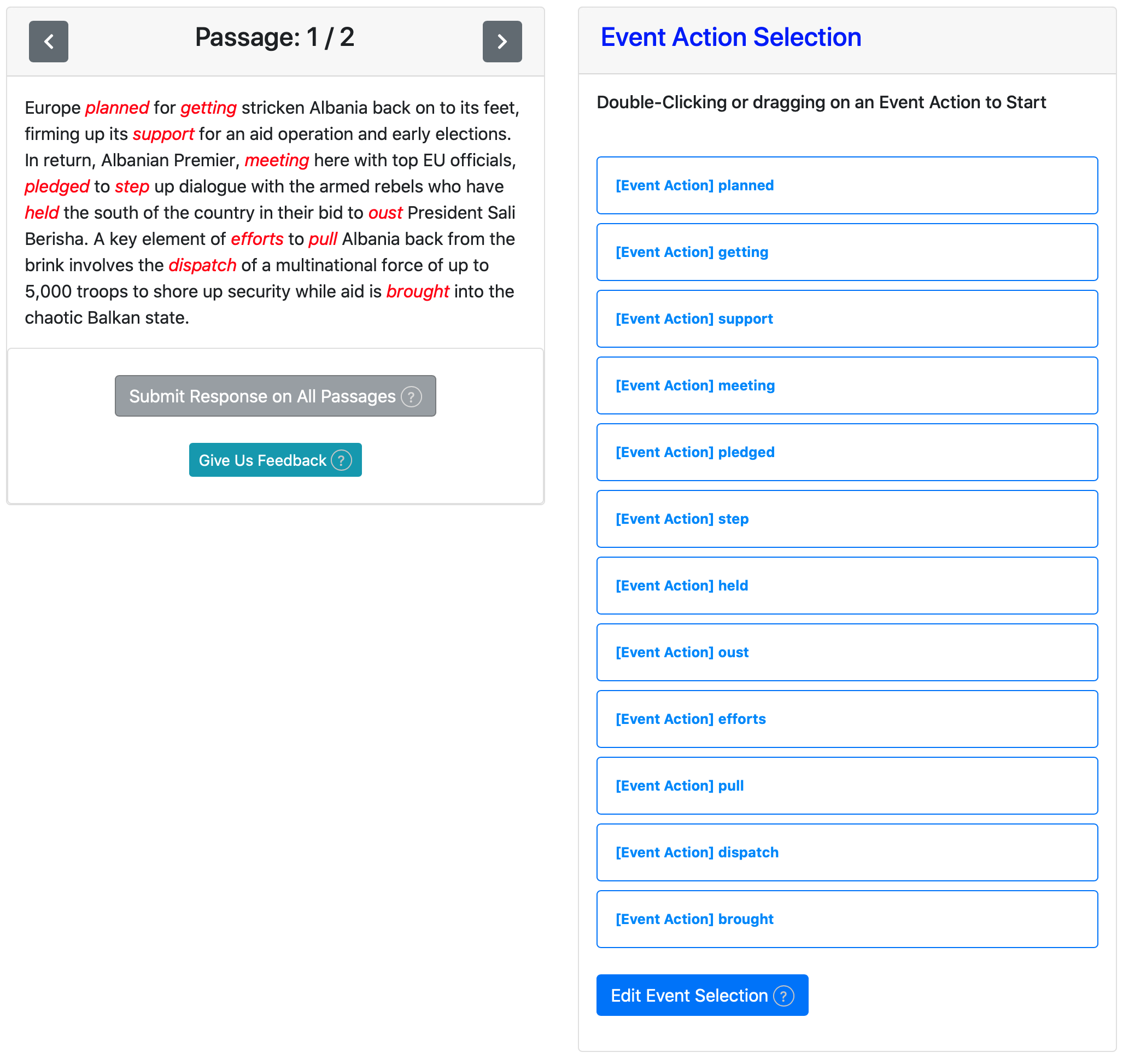}
    \caption{Event Selections Completed}
    \label{fig:interface-event2}
    \end{subfigure}
\caption{An Illustration of Event Selection Interface}
\label{fig:interface-event} 
\end{figure*}

\begin{figure*}[h!]
\centering
    \begin{subfigure}[b]{0.9\textwidth}
    \centering
    \includegraphics[width=\textwidth]{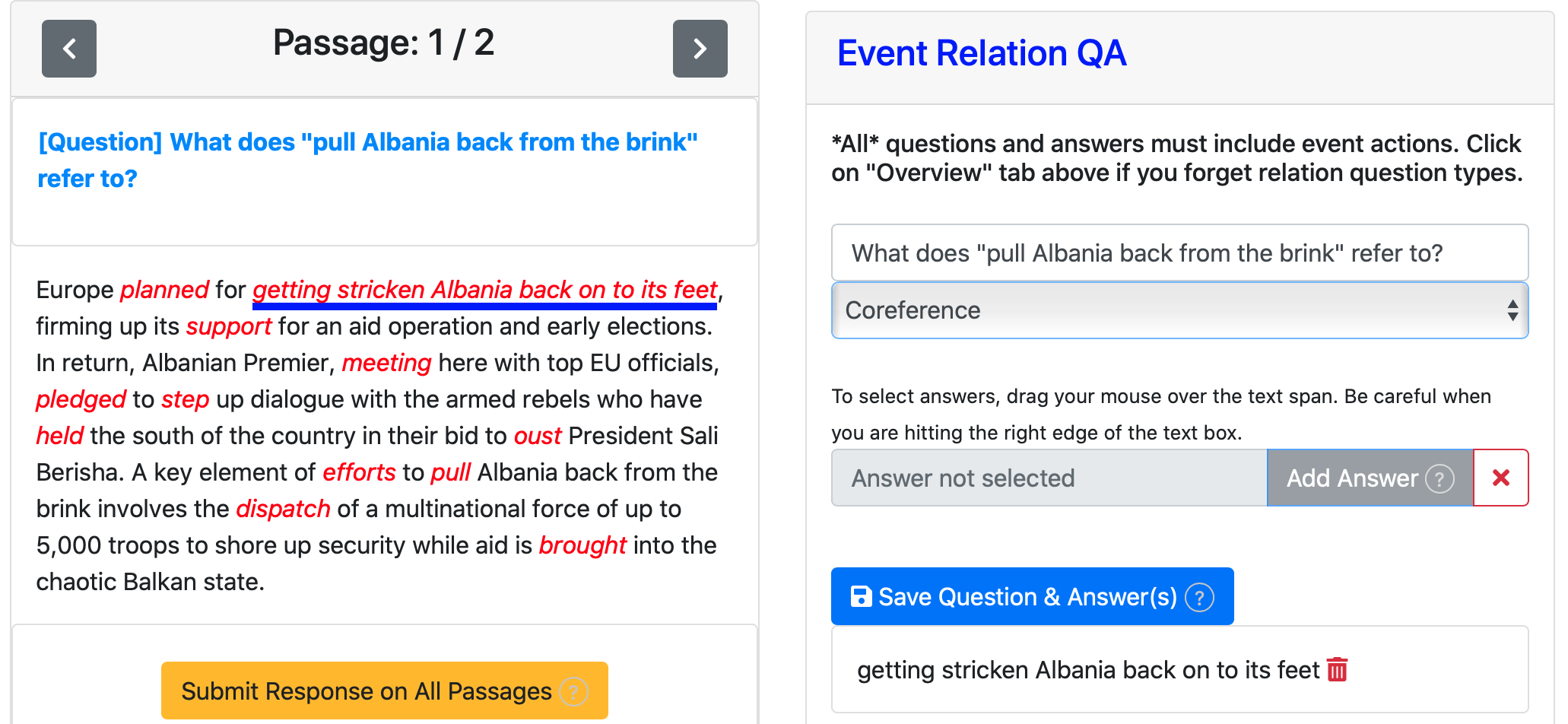}
    \caption{QA Annotations in Progress}
    \label{fig:interface-qa1}
    \end{subfigure}
    \par\bigskip
    \begin{subfigure}[b]{0.9\textwidth}
    \centering
    \includegraphics[width=\textwidth]{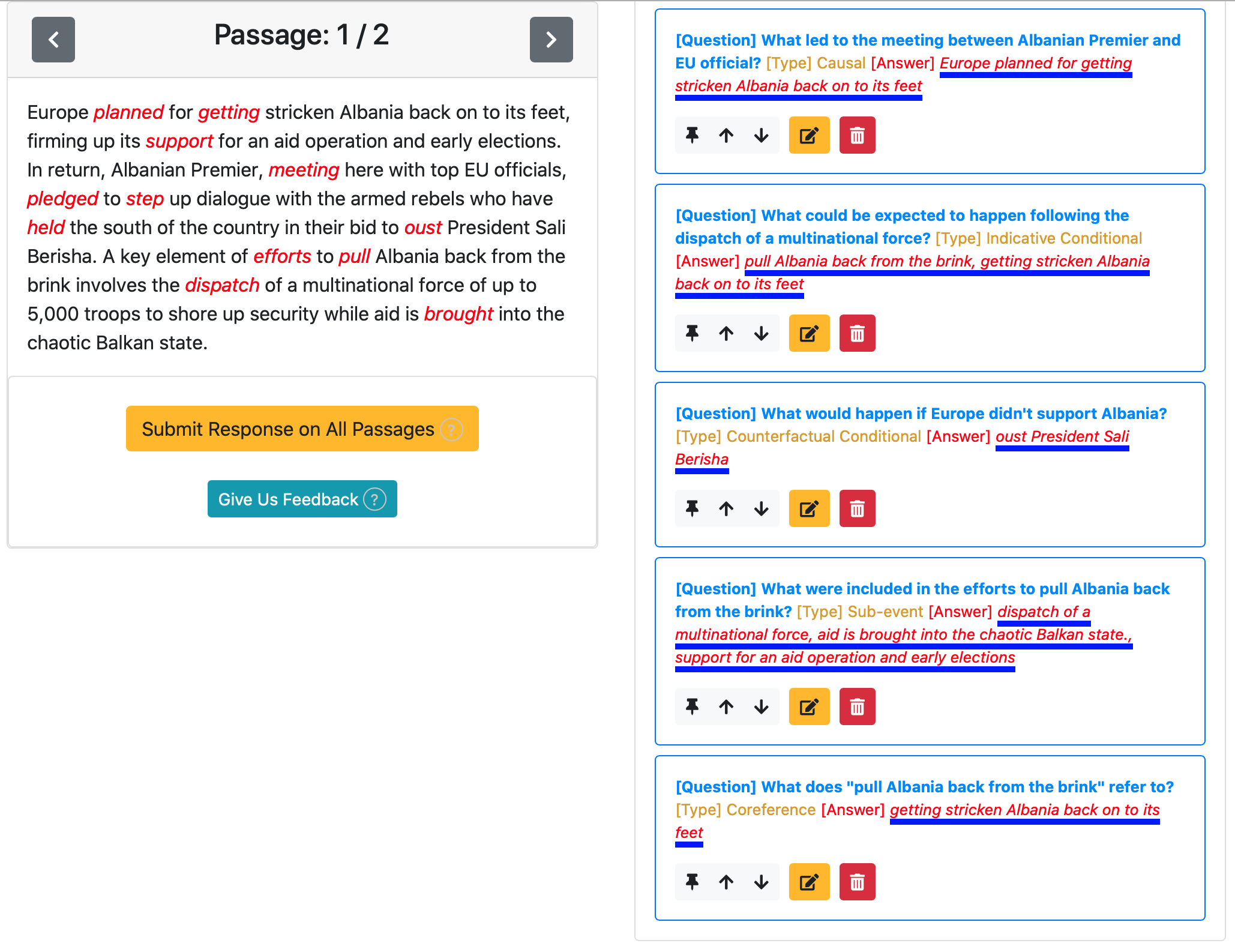}
    \caption{QA Annotations Completed}
    \label{fig:interface-qa2}
    \end{subfigure}
\caption{An Illustration of QA Interface}
\label{fig:interface-qa} 
\end{figure*}

Please refer to Figure~\ref{fig:interface-event} for user interface of event selection and Figure~\ref{fig:interface-qa} for QA annotation.

\section{QA validity}
A pair of QA is valid if and only if it fulfils the following criteria,
\begin{enumerate}
    \item Both questions and answers MUST contain correct events. Events in questions can have different textual form.
    \item Both questions and answers MUST be natural and meaningful. Workers with spotted spamming are immediately disqualified.
    \item The semantic relation formed by a QA pair MUST falls into one of the five relation categories we define.
\end{enumerate}

Note that QA validity is different from QA completeness for which we instruct workers to find all possible answers in the passage.

\section{Type Distribution}
\begin{figure}[h!]
\centering
    \includegraphics[width=0.95\columnwidth]{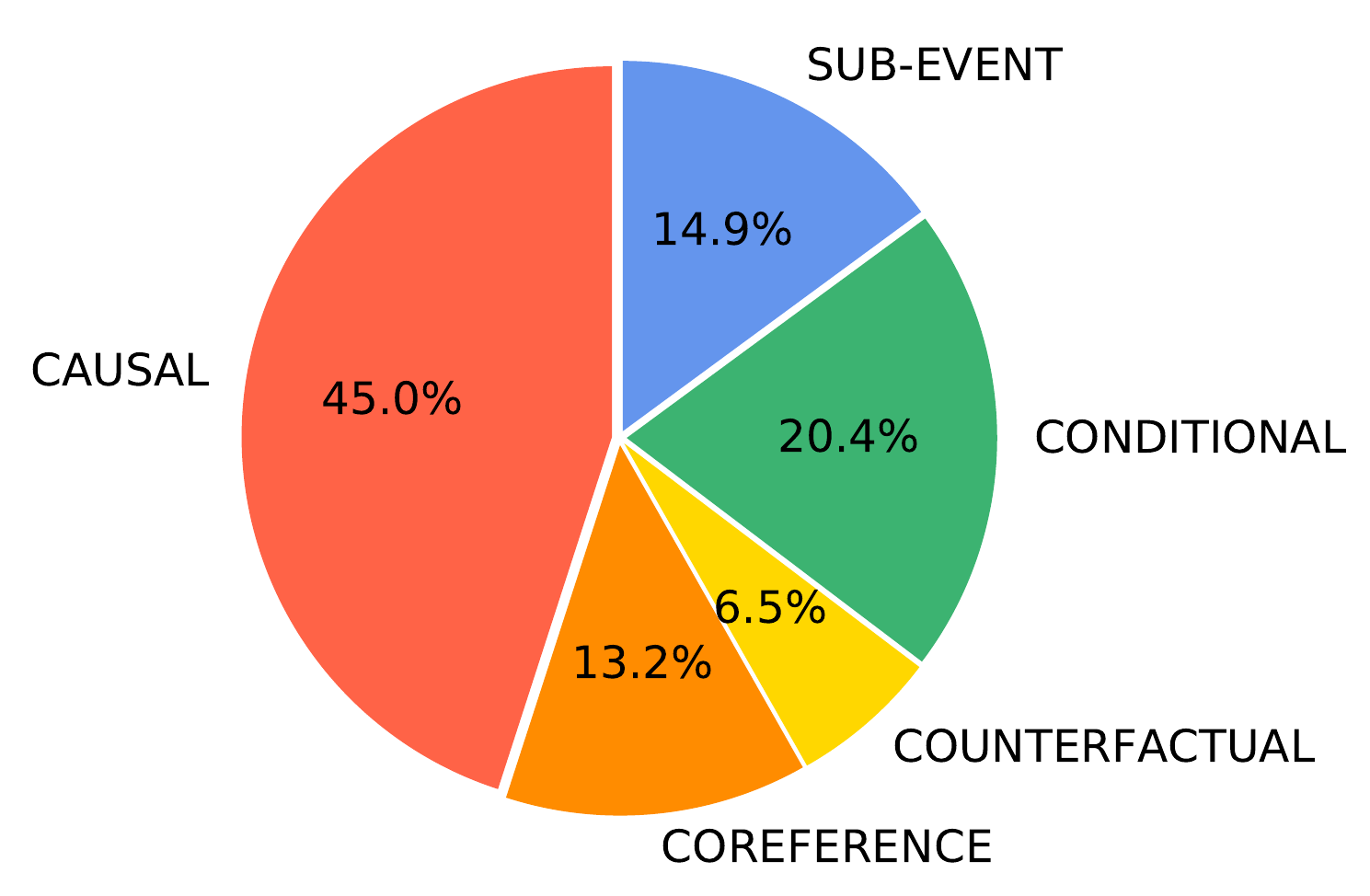}
    \caption{Type Distribution: train data}
\label{fig:type-dist-train} 
\end{figure}

\begin{figure}[h!]
\centering
    \includegraphics[width=0.95\columnwidth]{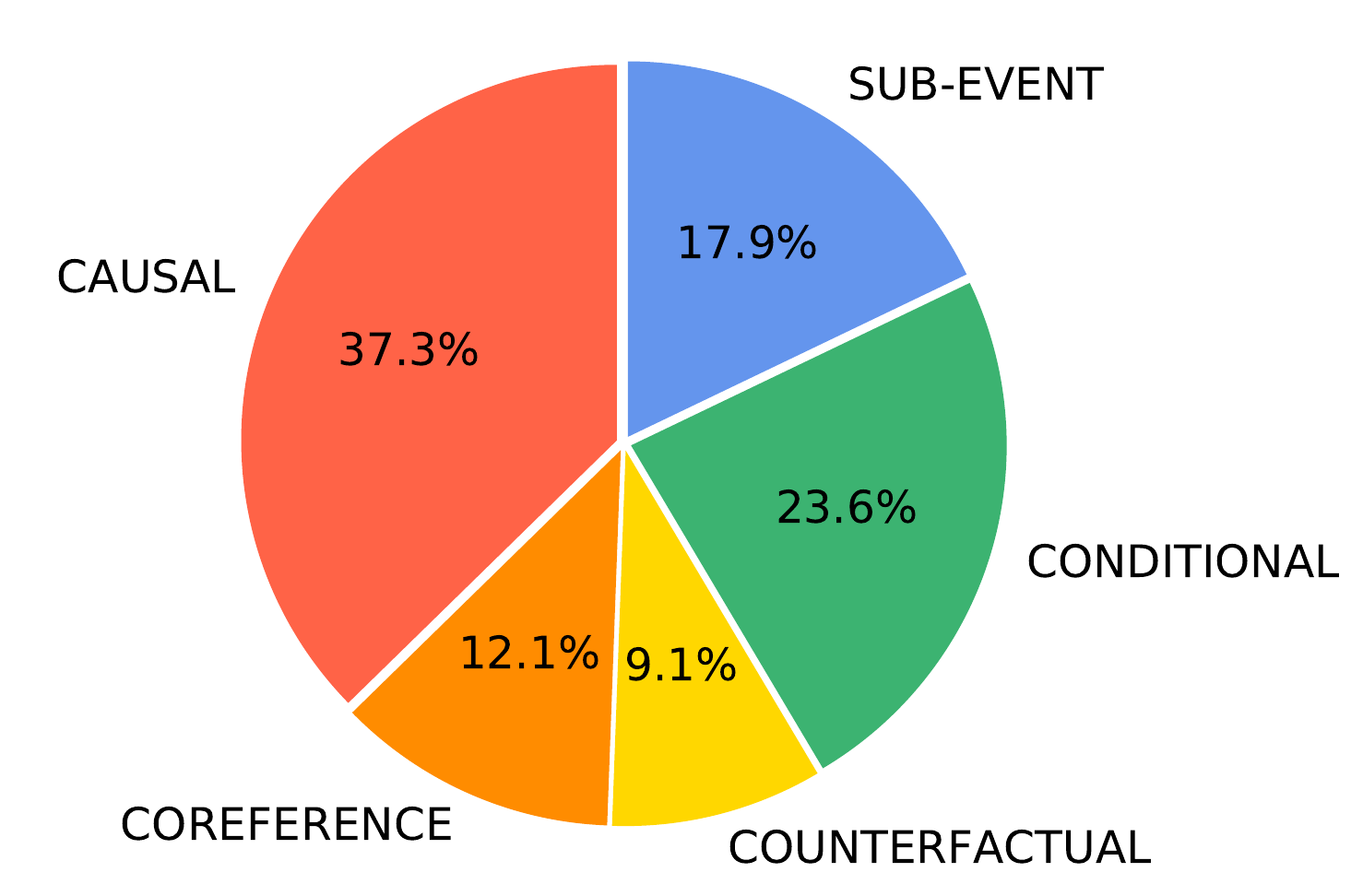}
    \caption{Type Distribution: evaluation data (dev + test)}
\label{fig:type-dist-dev} 
\end{figure}

Figure~\ref{fig:type-dist-train} \&~\ref{fig:type-dist-dev} compare the semantic relation type distribution between the train and evaluation data.

\paragraph{IAA Calculation.} A pair-wise micro F1 score is calculated by considering one of the annotations as ground truth and the other annotations as predictions. We then rotate the ground truth among all annotations for the same QA and take the average scores as the final IAA scores. For Fleiss's $\kappa$ score, we follow the same process described in the \citet{fleiss2013statistical} to evaluate the final IAA.

\section{Most Frequent N-grams}
\begin{figure}[h!]
    \centering
    \begin{subfigure}[b]{0.95\columnwidth}
        \centering
        \includegraphics[width=\columnwidth]{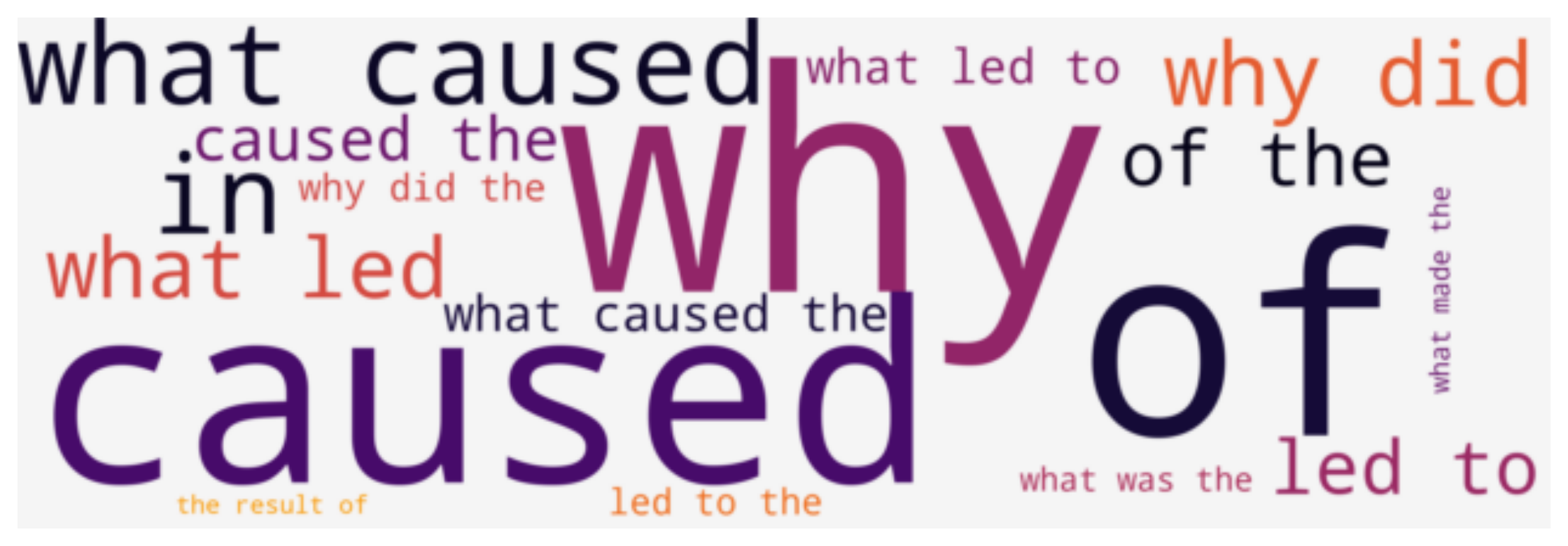}
        \caption{Most frequent n-grams for {\causal}}
        \label{fig:most_freq_causal} 
    \end{subfigure}
    \hfill
        \begin{subfigure}[b]{0.95\columnwidth}
        \centering
        \includegraphics[width=\columnwidth]{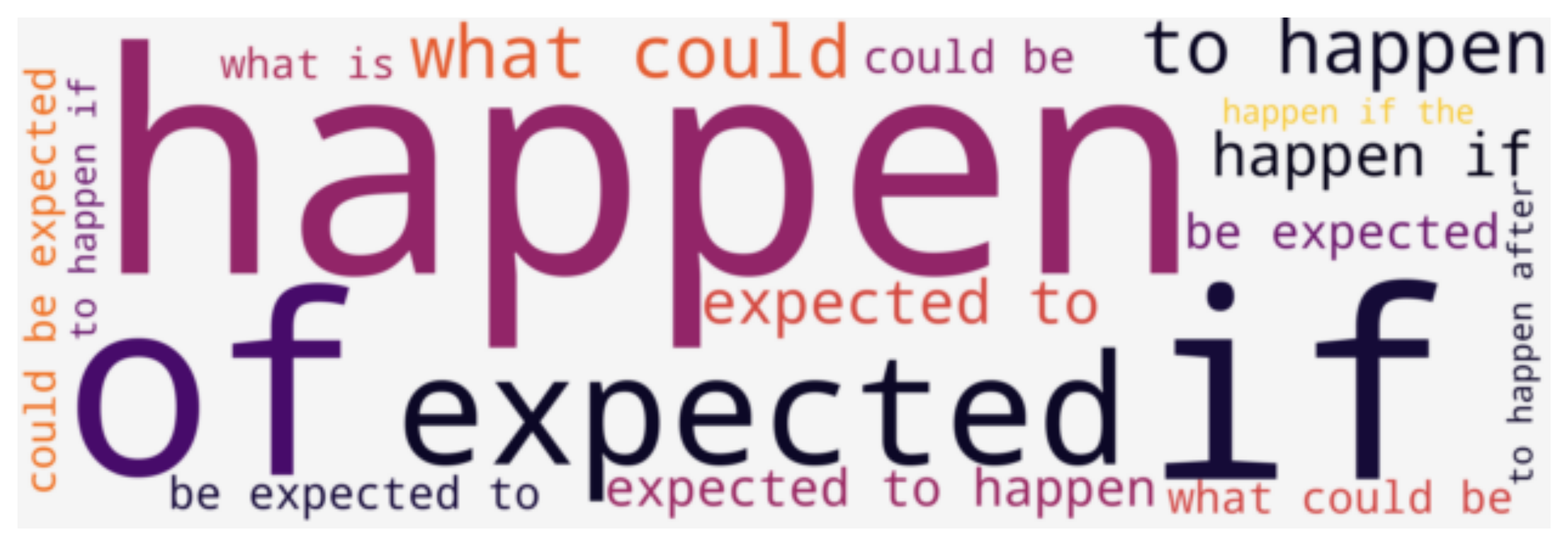}
        \caption{Most frequent n-grams for {\condition}}
        \label{fig:most_freq_condition} 
    \end{subfigure}
    \hfill
        \begin{subfigure}[b]{0.95\columnwidth}
        \centering
        \includegraphics[width=\columnwidth]{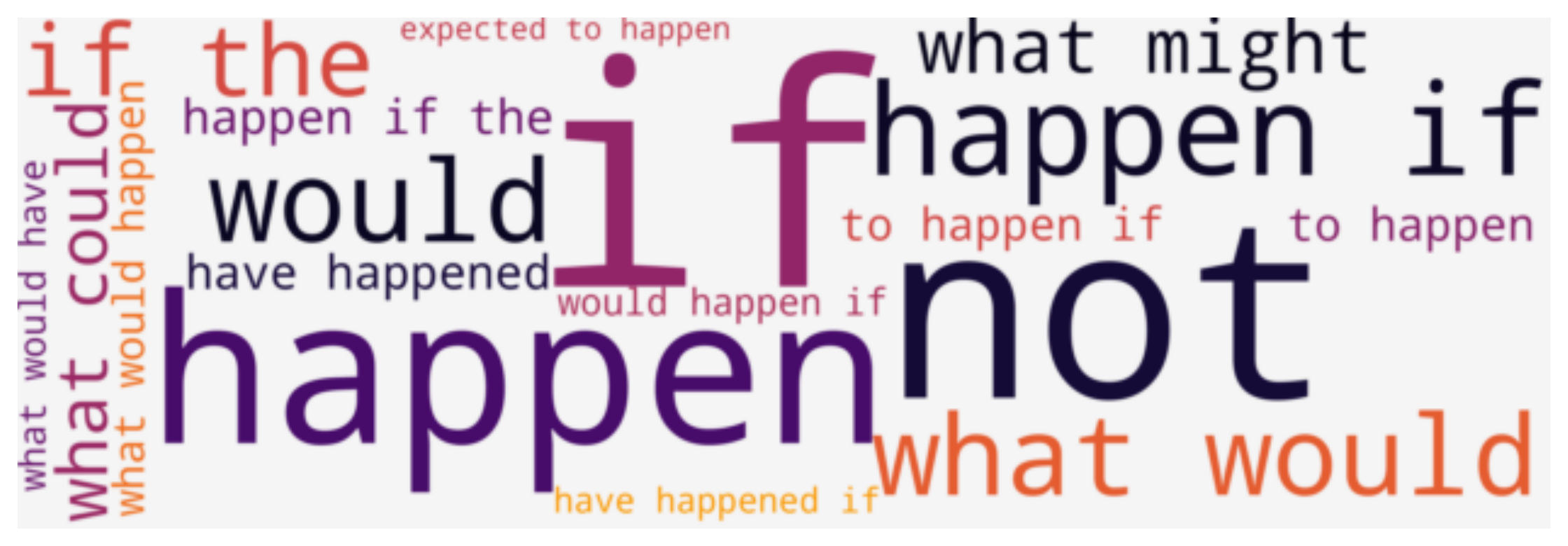}
        \caption{Most frequent n-grams for {\counter}}
        \label{fig:most_freq_counter} 
    \end{subfigure}
    \hfill
        \begin{subfigure}[b]{0.95\columnwidth}
        \centering
        \includegraphics[width=\columnwidth]{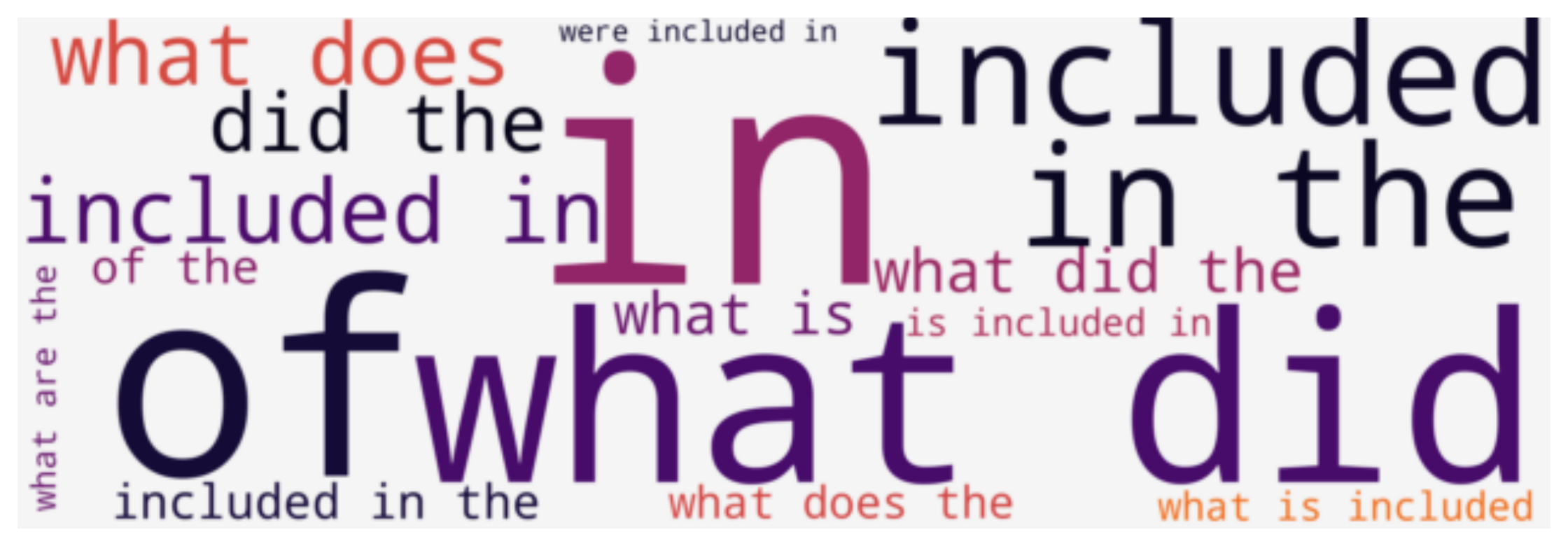}
        \caption{Most frequent n-grams for {\subevent}}
        \label{fig:most_freq_subevent} 
    \end{subfigure}
        \begin{subfigure}[b]{0.95\columnwidth}
        \centering
        \includegraphics[width=\columnwidth]{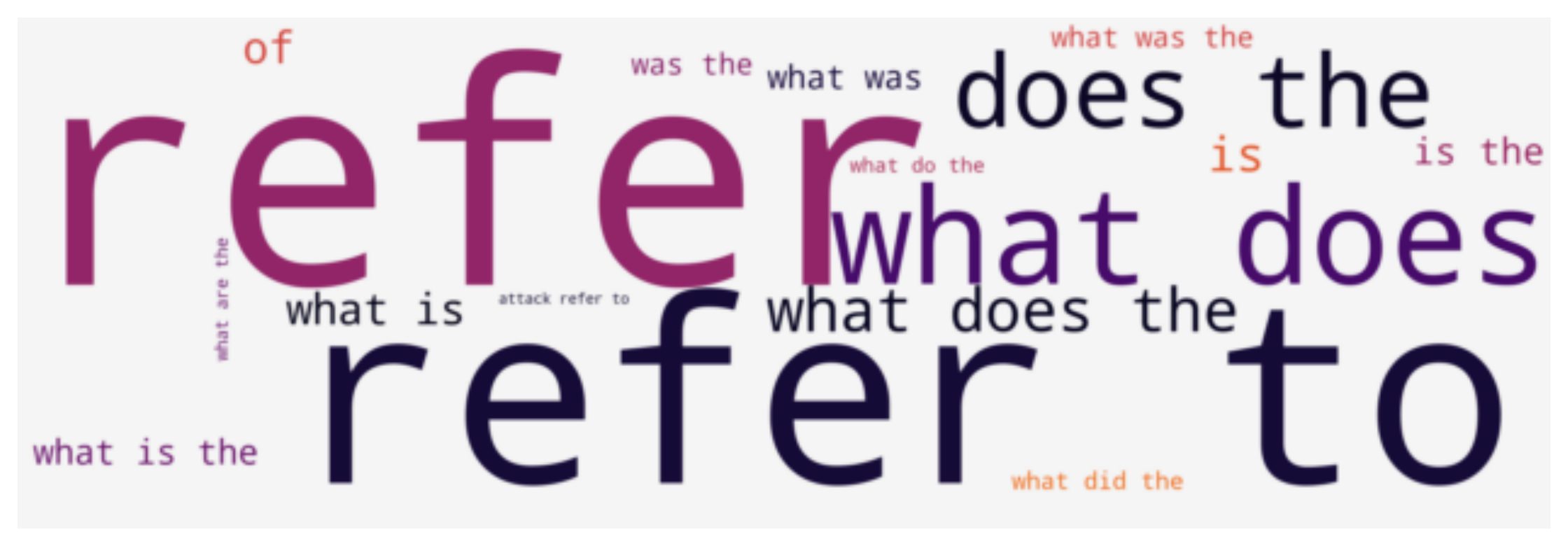}
        \caption{Most frequent n-grams for {\coref}}
    \label{fig:most_freq_coref} 
    \end{subfigure}
    \caption{Enlarged charts for most frequent n-grams in questions.}
    \vspace{-0.6cm}
\end{figure}

Enlarged imagines for frequent n-grams in questions can be found in Figure~\ref{fig:most_freq_causal}-\ref{fig:most_freq_coref}.

\begin{figure}[h!]
    \centering
\includegraphics[trim=0cm 0cm 0cm 0cm, clip, width=0.95\columnwidth]{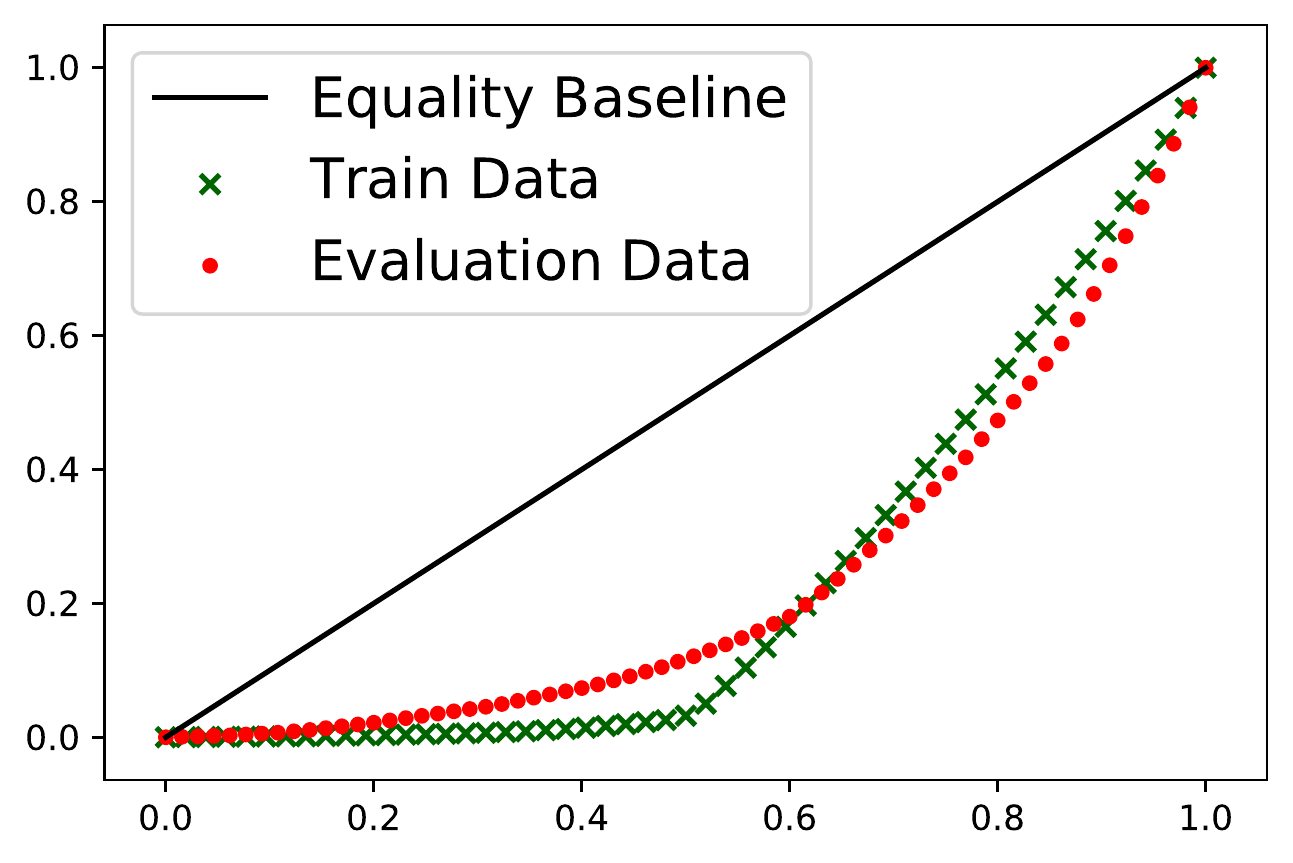}
\caption{Questions distributions by workers in train v.s. evaluation sets. Equality baseline indicates each participant provides equal number of questions.}
\label{fig:type-dist}
\end{figure}

\section{Worker Distribution}
\label{sec:worker-dist}
We had 70 workers in total who passed our qualification exam and completed at least 1 assignment in our project. Due to our rigorous validating process, only 27 were able to make it into Task 4 and the Large Task which consist of a large number of assignments. Figure~\ref{fig:type-dist}, known as Lorenze Curve \citep{lorenzcurve} illustrates the distribution of number of questions completed by workers. The equality baseline indicates the questions are perfectly well distributed among all workers, i.e. everyone completes the same numbers of questions. The further a curve deviates from the equality baseline, the more unevenly distributed a dataset becomes. Compared with the train data, we observe that the evaluation set is slightly better distributed, which reflects our validation process: for workers who failed our validation tasks and were disqualified, they could still provide some good quality QAs, which we keep in the evaluation data. This increases the diversity of the evaluation set.

\section{Number of Tokens.}
\label{sec:num-tokens}
Table~\ref{tab:ques-ans-tok-num} shows an average number of tokens in questions and answers. The {\counter} questions contain the most number of tokens as additional words are often needed to specify the negation reasoning. The average numbers of tokens are all around 6.5 across 5 types of answers. This is exactly the medium of our answer length limits where we set the minimum and maximum numbers of words to be 1 and 12 respectively. The average number of tokens in the passages is 128.1 with the longest passage containing 196 tokens.

\begin{table}[htbp!]
\centering
\scalebox{0.85}{
\setlength{\tabcolsep}{5pt}
\begin{tabular}{l|cc}
\toprule
& \multicolumn{2}{c}{\# Tokens}  \\
\midrule
Semantic Types & Question & Answer \\
\midrule
{\causal} & 10.3 & 6.6 \\
{\condition} & 12.1 & 6.4 \\
{\counter} & 13.7 & 6.0 \\
{\subevent} & 9.3 & 6.5  \\
{\coref} & 8.6 & 6.5 \\
\bottomrule
\end{tabular}
}
\caption{Average number of tokens in questions and answers.}
\label{tab:ques-ans-tok-num}
\end{table}

\section{Reproduction Check List}
We finetune BART-base, BART-large, T5-base, UnifiedQA-base and UnifiedQA-large on {\dataset}. UnifiedQA models are all based on T5. Hyper-parameters search ranges are 1) learning rate: $(1e^{-5}, 5e^{-5}, 1e^{-4})$; batch size: $(2, 4)$. Best hyper-parameters can be found in Table~\ref{tab:model-details}. We also use 3 random seeds: $(5, 7, 23)$ and report the average performances for each model. For RoBERTa-large, there is an additional hyper-parameter, positive token training weight mentioned in Section~\ref{sec:extractive-qa}, and it search range is $(1, 2, 5, 10, 20)$.

For BART-base, BART-large, T5-base and UnifiedQA-base models, we were able to finetune on a single Nvidia GTX2020 GPU with 11G memory. For Pegasus and UnifiedQA-large, we have to use a much larger Nvidia A100 GPU with 40G memory. We tried to finetune UnifiedQA based on T5-3B, but we were not able to fit batch size = 1 into a single Nvidia A100 GPU. So we stop at UnifiedQA-large. All reproduction details can be found in the separately submitted code.

\label{sec:model-details}
\begin{table}[h!]
\centering
\scalebox{0.75}{
\setlength{\tabcolsep}{5pt}
\begin{tabular}{l|ccc}
\toprule
Models & \# Params. & Best Hyper. & GPU \\
\midrule
RoBERTa-large & 355M & lr$=1e^{-5}$; b$=2$ & GTX2080\\
BART-base & 139M & lr$=5e^{-5}$; b$=4$ & GTX2080\\
BART-large & 406M & lr$=1e^{-5}$; b$=2$ & GTX2080\\
T5-base & 220M & lr$=1e^{-4}$; b$=4$ & GTX2080\\
UnifiedQA-base & 220M & lr$=5e^{-5}$; b$=2$ & GTX2080\\
UnifiedQA-large & 770M & lr$=5e^{-5}$; b$=4$ & A100\\
\bottomrule
\end{tabular}
}
\caption{Model and fine-tuning details. Learning rate: lr; batch size: b.}
\label{tab:model-details}
\vspace{-0.2cm}
\end{table}

\section{Sub-sample Performances}
\begin{figure}[h!]
    \centering
\includegraphics[trim=0cm 0cm 0cm 0cm, clip, width=0.95\columnwidth]{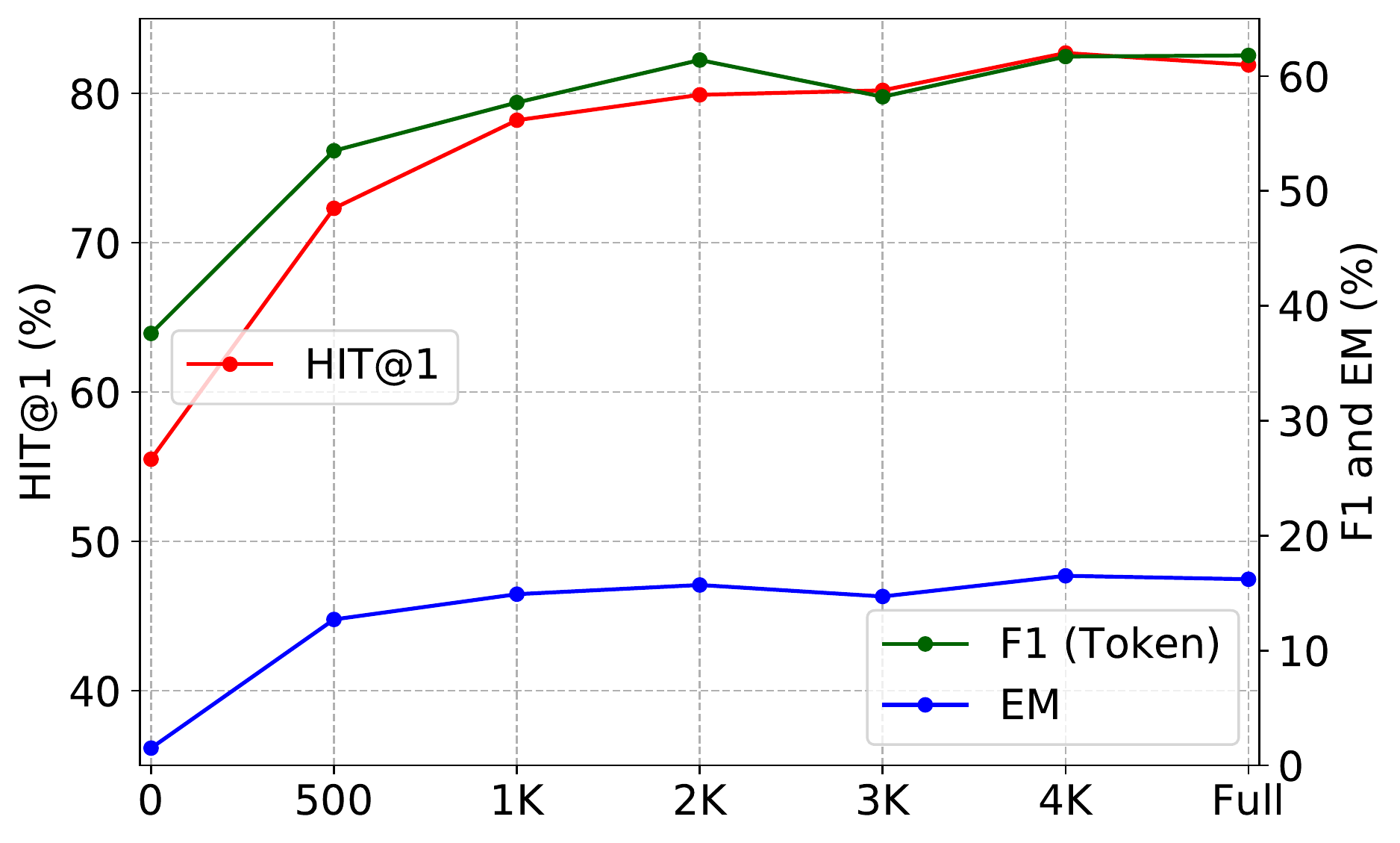}
\caption{Sub-sample fine-tuning performances for hierarchical relations: {\subevent} + {\coref}. All numbers are average over 3 random seeds.}
\label{fig:downsample-sub-coref}
\end{figure}

In Figure~\ref{fig:downsample-sub-coref} we show the fine-tuning UnifiedQA-large using different numbers of training samples for {\subevent} and {\coref}. We observe the same level-off after using 2K training data as in Figure~\ref{fig:downsample} for all semantic types.

\section{Model Significance Test}
\label{sec:significance-test}
To conduct statistical tests over model improvements, we pick the model with highest $F_1^T$ score among the three random seeds for the ``best'' hyper-parameters chosen in Table~\ref{tab:model-details}. We then perform McNemar's tests for \textbf{HIT@1} and \textbf{EM}. Specifically, if \textbf{HIT@1} = 1.0 for a sample, we treat it as a correct prediction; otherwise, it is incorrect. The same logic applies to \textbf{EM}. We only conduct statistical tests over pairs of models that are comparable in Table~\ref{tab:results-answer-gen}, and test results are shown in Table~\ref{tab:significance} below.

\begin{table}[h!]
\centering
\scalebox{0.8}{
\setlength{\tabcolsep}{5pt}
\begin{tabular}{l|cc}
\toprule
Model Comparisons & \textbf{HIT@1} & \textbf{EM} \\
\midrule
\textbf{Zero-shot} & &  \\
\midrule
T5-base $\rightarrow$ UnifiedQA-base & 0.790 & \ul{$\ll$0.001} \\
UnifiedQA-base $\rightarrow$ UnifiedQA-large & \ul{$\ll$0.001} & \ul{0.001} \\
\midrule
\textbf{Finetune} & &  \\
\midrule
BART-base $\rightarrow$ BART-large & 0.865 & 0.231 \\
T5-base $\rightarrow$ UnifiedQA-base & \ul{0.018} & 0.574 \\
UnifiedQA-base $\rightarrow$ UnifiedQA-large & \ul{$\ll$0.001} & \ul{0.022} \\
RoBERTa-large $\rightarrow$ UnifiedQA-base & \ul{$\ll$0.000} & \ul{0.001} \\
RoBERTa-large $\rightarrow$ UnifiedQA-large & \ul{$\ll$0.000} & \ul{$\ll$0.000}\\
\bottomrule
\end{tabular}
}
\caption{McNemar's test per \textbf{HIT@1} and \textbf{EM} metrics. Models on the right-hand side of ``$\rightarrow$'' are better. All numbers are p-values with $\le 0.05$ indicating statistically significant (underlined).}
\label{tab:significance}
\vspace{-0.2cm}
\end{table}

\section{Generative v.s. Extractive QA}
\label{sec:gen-vs-extract}
In Table~\ref{tab:ex-gen-extract}, we show 3 examples comparing predicted answers between generative and extractive QA. In general, scattered answers occur frequently in extractive QA, but barely occur in generative QA. In other words, generative QA is able to consistently produce complete and meaningful answers.
\begin{table*}[h!]
\centering
\scalebox{0.95}{
\begin{tabular}{l}
\toprule
\textbf{Ex. 1} \\
\ul{\textbf{Passage:}} \hl{the} ser\hl{bs} only lifted their threat of a boycott friday after heavy international pressure and \\ the intervention of serbian president slobodan milosevic, a longtime supporter of the rebels. \\
in a last-minute attempt to get people to vote, the independent democratic serb party (sdss), led by \\ vojislav stanimirovic, launched into what seemed more like a mobilisation rather than a real \\ political campaign. \\
\ul{\textbf{Question:}} what could happen if there was no intervention by the serbian president?\\
\ul{\textbf{Generative Answers:}} 1. a boycott \\
\ul{\textbf{Extractive Answers:}} 1. \hl{the}; 2. \hl{bs}; 3. lifted their threat of a boycott \\
\midrule
\textbf{Ex.2} \\
\ul{\textbf{Passage:}} french defence minister michele alliot-marie on sunday stressed paris's support for the \\ government of lebanese prime minister fuad siniora during a visit to the crisis-wracked nation.
\\ "i have come to reaffirm france's support for the legitimate government of lebanon," she told \\ reporters after meeting her lebanese counterpart elias murr.
alliot-marie also stressed paris's \\ backing for the beirut government to "exercise its sovereignty completely", and that the lebanese \\ army play "a role across all its territory".
le\hl{banon is undergoing a political crisis} with opposition \\ led by shiite movement hezbollah seeking \hl{to} bring down siniora's government and install a \\ government of national unity.
the french minister, who arrived in beirut on saturday for a 48-hour \\ visit, was also to meet siniora before heading to south lebanon for new year's eve with the french \\contingent of the united nations interim force in lebanon (unifil). \\
\ul{\textbf{Question:}} what caused alliot-marie to visit lebanon? \\
\ul{\textbf{Generative Answers:}} 1. lebanon is undergoing a political crisis \\
\ul{\textbf{Extractive Answers:}} 1. \hl{banon is undergoing a political crisis}; 2. \hl{to}; 3. bring down \\ siniora's government \\
\midrule
\textbf{Ex.3} \\
\ul{\textbf{Passage:}} vieira seems very enthusiastic about bringing in chinese capital and technology into the \\ west african country. he said priorities for bilateral cooperation could \hl{expand to} ports, roads, \\ bridges and mineral resources.
inspired by vieira's enthusiasm, cmec vice president zhou li promised\\ that \hl{a} special team would fly to guinea-bissau to discuss the details.
vieira reminded her that apart \\from guinea-bissau, other west african countries such as senegal and guinea also need power- \\generation facilities badly.
regarding china as a strategic friend who offers aids without political \\strings, many african countries impressed with the country's two-digit economic growth are seizing \\time to \hl{explore} cooperative opportunities during their stay in beijing to boost domestic economy.\\
\ul{\textbf{Question:}} what does the bilateral cooperation include?\\
\ul{\textbf{Generative Answers:}} 1. bringing in chinese capital and technology; 2. a special team would fly to \\ guinea-bissau; 3. talk about the details; 4. explore cooperative opportunities during their stay \\ in beijing\\
\ul{\textbf{Extractive Answers:}} 1. bringing in chinese capital and technology; 2. \hl{expand to}; \\ 3. ports, roads, bridges and mineral resources; 4. \hl{a}; 5. special team would fly to guinea-bissau; \\ 6. discuss the details; 7. \hl{explore}\\
\bottomrule
\end{tabular}
}
\caption{Examples of answers predicted by generative v.s. extractive QA models. Some passages are shortened for demonstration purpose. Incomplete predictions from extractive QA are highlighted.}
\label{tab:ex-gen-extract}
\vspace{-0.2cm}
\end{table*}

\end{document}